\PassOptionsToPackage{hidelinks}{hyperref}
\documentclass[journal]{IEEEtran}
\usepackage{amsmath,amsfonts}
\usepackage{graphicx}
\usepackage{cite}
\usepackage{bbm}
\usepackage{multirow}
\usepackage{xspace}
\usepackage{booktabs}
\usepackage{orcidlink}
\newcommand{\ie}{{\emph{i.e.}},\xspace}

\usepackage{bbding}

\usepackage{array}
\usepackage{threeparttable}

\newcommand{\tablesmallfont}{\fontsize{7.0}{9.5}\selectfont}

\begin{document}

\title{Unsupervised Collaborative Domain Adaptation \\ for Driving Scene Parsing}

\author{
Jiahe~Fan$^{\orcidlink{0000-0002-2364-6811}\,}$,~\IEEEmembership{Student Member,~IEEE},
Shaolong~Shu$^{\orcidlink{0000-0002-6759-2894}\,}$,~\IEEEmembership{Senior Member,~IEEE},
Mingjian~Sun$^{\orcidlink{0000-0001-8719-524X}\,}$,Tiehua Zhang$^{\orcidlink{0000-0002-7195-4472}\,}$,\\
Bohong~Xiao$^{\orcidlink{0009-0008-6153-2605}\,}$,
Hanli~Wang$^{\orcidlink{0000-0002-9999-4871}\,}$,~\IEEEmembership{Senior Member,~IEEE},
Rui Fan$^{\orcidlink{0000-0003-2593-6596}\,}$,~\IEEEmembership{Senior Member,~IEEE}

\thanks{Jiahe Fan, Shaolong Shu, and Rui Fan are with the College of Electronic and Information Engineering, Tongji University, Shanghai 201804, China (e-mails: \{jiahe\_fan, shushaolong, rfan\}@tongji.edu.cn).}
\thanks{Mingjian Sun is with the Department of Control Science and Engineering, the Harbin Institute of Technology, Harbin, Heilongjiang, 150001, China (e-mail: sunmingjian@hit.edu.cn).}
\thanks{Bohong Xiao is with the Department of Vehicle Control System and Software Development, NIO 201804, Shanghai, China. (e-mail: bohong.xiao@nio.com).}
\thanks{Tiehua Zhang is with the School of Computer Science and Technology, Tongji University, Shanghai, 201804, China (e-mail: tiehuaz@tongji.edu.cn).
}
\thanks{Hanli Wang is with the College of Electronic and Information Engineering, the School of Computer Science and Technology, and the Key Laboratory of Embedded System and Service Computing (Ministry of Education), Tongji University, Shanghai 201804, China (e-mail: hanliwang@tongji.edu.cn).}
}


\maketitle

\begin{abstract}

Reliable driving scene parsing is a fundamental capability for autonomous vehicles operating in open and dynamic driving environments.
However, adapting perception models to new deployment domains remains challenging because pixel-level annotations are expensive to obtain, while source-domain data are often inaccessible due to privacy, security, or ownership constraints.
Existing source-free unsupervised domain adaptation methods typically rely on a single pre-trained source model, which makes the adapted perception system vulnerable to source-specific biases and limits its robustness under diverse road layouts, illumination conditions, weather patterns, and traffic conditions.
This article presents an unsupervised collaborative domain adaptation (UCDA) framework for driving scene parsing in a source-free setting, which transfers complementary knowledge from multiple pre-trained source models to a unified target model without accessing any original source samples.
To compare predictions from independently trained models, UCDA constructs a class-level prototype memory bank and estimates cross-model prediction reliability through prototype similarity, reducing the effect of inconsistent confidence scales across source models.
Based on the resulting complementary supervision, UCDA adopts a two-stage transfer strategy: multiple source models are first refined on unlabeled target-domain driving data through collaborative optimization with positive and negative consistency constraints, and their validated expertise is then distilled into a single deployable target model.
Comprehensive evaluations on public driving-scene datasets and real-world data collected from an autonomous vehicle platform demonstrate that UCDA effectively consolidates complementary multi-source knowledge, improving target-domain scene parsing reliability and generalization across diverse driving environments.

\end{abstract}

\begin{IEEEkeywords}
Driving scene parsing, source-free domain adaptation, multi-source knowledge transfer, prototype-based reliability estimation
\end{IEEEkeywords}

\section{Introduction}
\label{sec.introduction}

\begin{figure*}[t!]
    \centering
    \includegraphics[width=0.9999999\textwidth]{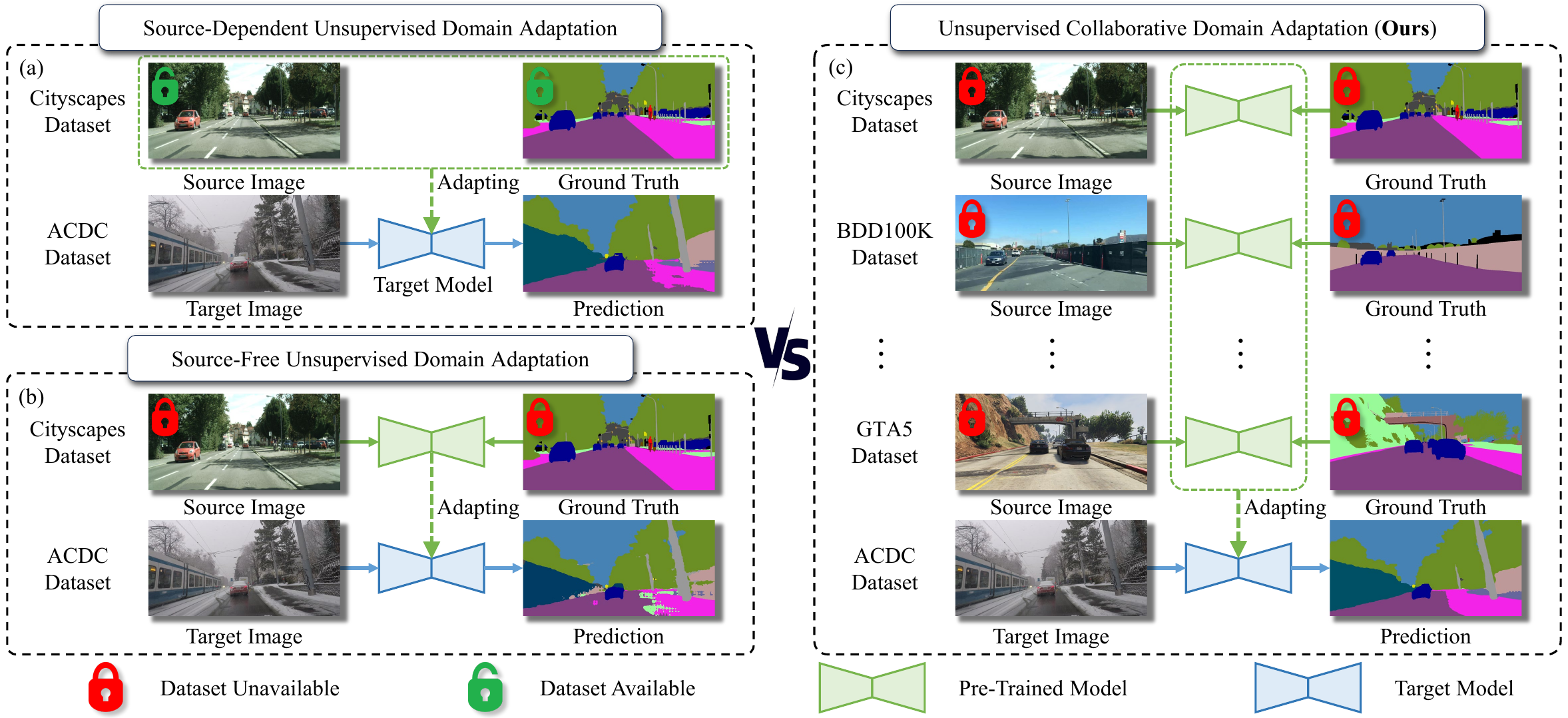}
    \caption{An illustration of (a) source-dependent and (b) source-free unsupervised domain adaptation paradigms, in comparison with the proposed (c) \textbf{unsupervised collaborative domain adaptation} paradigm. The proposed adaptation paradigm enables the transfer of prior knowledge from multiple source-domain models to the target model, significantly enhancing driving scene parsing performance compared to both source-dependent and source-free paradigms.}
    \label{fig.background}
\end{figure*}

\IEEEPARstart{R}{eliable} environmental perception is essential for autonomous vehicles operating in open and diverse urban environments.
Among perception tasks, driving scene parsing provides pixel-level semantic understanding of the surrounding environment, supporting downstream autonomy modules such as motion planning~\cite{Motion_Planning}, semantic occupancy estimation~\cite{OccNeRF}, and behavior prediction~\cite{ZhangMotions}.
Although deep learning models have substantially advanced driving scene parsing, their performance often degrades after deployment in new operating environments, such as unseen cities, weather conditions, illumination patterns, or sensor configurations~\cite{DALI}.
Adapting perception models to these deployment environments remains challenging, since collecting and annotating large-scale driving data for each new environment is labor intensive and costly~\cite{TGU}.
Moreover, raw driving images captured by autonomous vehicles in public environments may contain privacy-sensitive information, such as identifiable pedestrians and vehicle license plates, and are often constrained by regulatory, institutional, or commercial restrictions~\cite{sfdass,csfda}.
To reduce the need for annotated data in each new deployment environment, unsupervised domain adaptation (UDA) has been extensively explored to transfer knowledge from annotated data collected in existing environments (source domains) to unlabeled data collected in new environments (target domains)~\cite{caco,aras,pbal,wang2024curriculum,Tiandbc,Caoiapc,SND}.
Depending on whether the original source-domain data remain accessible during adaptation, UDA methods can be broadly categorized into two paradigms: source-dependent and source-free UDA~\cite{HG-SFDA}.
As illustrated in Fig.~\ref{fig.background}(a), source-dependent UDA approaches~\cite{caco,aras,pbal} require access to the original source training data during adaptation, which is often infeasible for autonomous driving applications due to privacy regulations, data security risks, and intellectual property restrictions~\cite{ucs}.
In contrast, source-free UDA (SFUDA) approaches~\cite{SFDASEG,Zhou_2025,CoUDA} have emerged as a practical alternative for privacy-preserving adaptation.
As illustrated in Fig.~\ref{fig.background}(b), this paradigm adapts a pre-trained source model to the target domain using only unlabeled data from the deployment environment, thereby eliminating source-data dependency and its inherent privacy risks~\cite{Deng_2025_CVPR}.
Consequently, this paradigm has gained significant attention as data sharing becomes increasingly restricted across organizations and regions.

Nevertheless, relying on a model pre-trained on a single source-domain dataset inherently constrains the generalizability of UDA approaches, as it often tends to overfit to domain-specific characteristics rather than learning robust and transferable representations~\cite{rsd}.
This limitation becomes particularly critical for autonomous driving systems, where source and target domains may differ substantially in visual appearance, road layout, weather, illumination, and traffic composition~\cite{CLIPP}.
For instance, the target-domain datasets may involve urban layouts, adverse weather, nighttime illumination, or sensor-specific characteristics that are insufficiently represented in the source data.
Moreover, discrepancies in semantic class distributions, especially the under-representation of safety-critical or long-tail categories in the source domain, can further hinder effective knowledge transfer~\cite{ama}.
These challenges motivate the use of complementary expertise from multiple pre-trained source models to overcome the inherent limitations of single-source UDA in scalable autonomous driving perception.

However, reliably transferring complementary expertise from multiple pre-trained source models to a target model remains challenging~\cite{lmam}.
Since source models are typically trained on data collected from different source domains, such as different cities, weather conditions, sensor configurations, or traffic scenarios, they tend to develop distinct perception strengths.
For example, one model may provide more reliable parsing in clear daytime scenes, whereas others may be more robust in tunnels, nighttime scenes, or adverse weather conditions~\cite{simt}.
Such specialization may also appear across semantic categories: certain models may better capture dominant road-scene structures such as roads and buildings, whereas others may be more effective for safety-critical or long-tail objects such as riders and traffic lights~\cite{aan}.
Without ground-truth annotations in the target domain, it is difficult to determine which source model provides the most trustworthy prediction for each scene, category, or image region.
Standard confidence indicators, such as softmax probability~\cite{calsfda} and entropy~\cite{Vu_2019_CVPR}, are often poorly calibrated across independently trained models.
Consequently, different source models may produce confidence scores on incompatible scales, causing unreliable predictions to be selected as supervision~\cite{wang2024curriculum}.
This reliability mismatch hinders the effective integration of complementary expertise from multiple source models for robust autonomous driving perception.

Beyond the difficulty of aligning inconsistent predictions, another critical challenge lies in how complementary expertise can be extracted and utilized from multiple source models in complex driving environments.
Several recent approaches~\cite{manf,msud,ZHAO20251} use high-confidence predictions as pseudo labels to guide the target model.
While such a strategy provides initial guidance for adaptation, it overlooks the rich structural information in uncertain or low-confidence regions~\cite{wang2024curriculum}.
This limitation is particularly important for perception in autonomous driving, where different source models may share common failure modes under rare or difficult operating conditions.
For example, all source models may become uncertain when encountering unusual traffic layouts, degraded visibility, or long-tail objects that are insufficiently represented in their training data~\cite{mdaf}.
In such cases, relying only on high-confidence predictions prevents the adaptation process from exploiting informative cues in difficult regions and may reinforce biased or incorrect predictions.
This can lead to negative transfer, where errors are repeatedly propagated during adaptation.
Therefore, an effective collaborative adaptation framework should exploit both reliable and uncertain predictions, using confident regions for explicit semantic guidance while extracting complementary supervisory cues from ambiguous regions to mitigate shared model weaknesses.

To address these challenges, this article presents an unsupervised collaborative domain adaptation (UCDA) framework for driving scene parsing.
As illustrated in Fig.~\ref{fig.background}(c), UCDA exploits multiple models pre-trained on diverse source-domain data and collaboratively transfers their complementary knowledge to the target model without accessing the original source data.
To this end, the framework first establishes a unified criterion for evaluating prediction reliability across different source models, semantic categories, and image regions.
This reliability assessment reduces inconsistencies caused by model-specific confidence scales, which is critical for selecting trustworthy supervisory signals in the unlabeled target domain.
Specifically, the framework constructs class-level prototypes for each source model by computing logit centroids from regions where multiple source models reach high-confidence consensus.
These prototypes serve as semantic anchors for measuring the alignment between pixel-level predictions and class-level representations.

As a result, prediction reliability is estimated based on the consistency between pixel-level logits and class-level logit prototypes rather than raw confidence scores, enabling more reliable comparison across independently trained source models.
The resulting reliability metric identifies model-specific perception strengths that are distributed across source domains.
Although integrating these complementary strengths can improve scene parsing in the target driving environment, directly combining multiple source models may introduce interference because each model still carries source-specific biases.
To address this issue, UCDA adopts a two-stage adaptation strategy that separates source-model refinement from final target-model distillation.
In the first stage, the source models are collaboratively refined using unlabeled data from the target driving environment, enabling them to exchange complementary knowledge and reduce individual knowledge gaps in the target domain.
The refined source models then serve as teachers in the second stage, where their complementary expertise is distilled into a unified target model for independent deployment.
This design allows final distillation to rely on refined complementary knowledge rather than directly aggregating biased predictions from source models on target-domain data.
Collaborative refinement is further implemented through a decoupled learning paradigm with positive and negative consistency constraints.
High-confidence regions provide explicit semantic guidance through refined pseudo labels, whereas uncertain regions are regulated by dual consistency signals that reinforce likely categories and suppress unlikely class assignments.
This mechanism helps resolve semantic ambiguities in low-confidence regions, which is important for robust scene parsing in complex driving environments.
The proposed framework is evaluated on diverse cross-domain driving scene parsing tasks, covering both synthetic-to-real and real-to-real adaptation scenarios.
In addition to evaluations on public datasets, UCDA is further tested on data collected from an autonomous vehicle platform to examine its practical applicability in real-world driving environments.
Experimental results demonstrate that UCDA improves scene parsing reliability across complex urban driving environments compared with existing UDA methods.

In a nutshell, this study makes the following contributions:

\begin{itemize}

%
%
%

\item A novel unsupervised collaborative domain adaptation framework that integrates complementary knowledge from multiple pre-trained source models for driving scene parsing without accessing the original source data.

\item A prototype-aware cross-model reliability estimation mechanism that mitigates model-specific confidence discrepancies and enables trustworthy supervision selection on unlabeled target-domain data.

\item A two-stage knowledge transfer strategy that refines source models through collaborative optimization and distills their validated expertise into a unified target model.

\item Extensive evaluations on synthetic-to-real and real-to-real adaptation tasks, together with real-world driving data collected from an autonomous vehicle platform, demonstrating improved robustness in complex urban driving environments.

\end{itemize}

The remainder of this article is structured as follows: Sect.~\ref{sec.related_works} reviews related SoTA UDA approaches. Sect.~\ref{sec.methodology} details the proposed UCDA framework.
Sect.~\ref{sec.exp} presents both quantitative and qualitative experimental results and compares the proposed approach with other SoTA UDA approaches.
Sect.~\ref{sec.discussion} discusses the limitations of the proposed approach while highlighting its potential applications. Finally, Sect.~\ref{sec.conclusion} concludes the article and discusses potential directions for future research.

\section{Related Work}
\label{sec.related_works}

\subsection{Source-Free Unsupervised Domain Adaptation}
\label{sec.related_works.sfuda_for_semantic_segmentation}

Source-free UDA approaches adapt a pre-trained source model to a target domain without accessing the original source data, mainly through pseudo source data generation~\cite{sfdass,vdmda,aan} or self-training with pseudo labels~\cite{urma,gsdaa,calsfda,csfda}.
Generation-based methods synthesize source-like images or feature representations to recover missing source-domain information and support conventional distribution alignment~\cite{sfdass}.
However, these methods often introduce substantial computational overhead due to the complexity of image or feature synthesis~\cite{vdmda}.
They may also suffer from distribution mismatch when synthesized samples fail to represent the characteristics of the original source domain~\cite{aan}.
Self-training methods generate supervisory signals from target-domain predictions through entropy minimization or consistency regularization~\cite{gsdaa}.
However, they are prone to error accumulation, as incorrect pseudo labels can be repeatedly reinforced during iterative adaptation.
This problem is further aggravated by poorly calibrated confidence scores, which make it difficult to distinguish reliable predictions from noisy ones on unlabeled target-domain data~\cite{calsfda}.
To address these limitations, UCDA exploits collaborative knowledge from multiple source models and introduces dual-signal supervision to improve prediction reliability estimation and reduce error accumulation caused by individual model biases.

\subsection{Multi-Source Knowledge Transfer in Domain Adaptation}
\label{sec.related_works.msk}

Multi-source knowledge transfer has been widely explored in UDA to alleviate the bias introduced by relying on a single source domain~\cite{zhou2024cycle,zhao2019multi,Xu_2018_CVPR}.
Conventional multi-source UDA methods assume access to labeled data from multiple source domains during adaptation, and learn transferable representations through distribution alignment~\cite{peng2019moment}, source weighting~\cite{Xu_2018_CVPR}, or source-specific knowledge aggregation~\cite{zhou2024cycle}.
Despite their advantages over single-source adaptation, their reliance on original source data limits their use in privacy-sensitive scenarios where images and annotations cannot be shared.
To relax this requirement, recent studies on image classification have explored multi-source source-free domain adaptation, where only multiple pre-trained source models and unlabeled target data are available. Representative methods combine source hypotheses~\cite{ahmed2021unsupervised}, construct confident anchors for pseudo-labeling~\cite{dong2021confident}, estimate source importance through discriminability and transferability~\cite{han2023discriminability}, or analyze source aggregation from the bias-variance perspective~\cite{shen2023balancing}. However, their reliability estimation is mainly performed at the image, sample, or domain level.
For dense prediction tasks, multi-source UDA has also been investigated in semantic segmentation, where multiple labeled source domains are exploited to improve target-domain scene parsing~\cite{zhao2019multi,he2021multi}. However, these methods generally require access to source-domain data during adaptation.
In contrast, existing multi-source source-free studies mainly focus on image classification, leaving reliability-aware knowledge transfer for driving scene parsing in a source-free setting insufficiently explored.
To address this gap, UCDA explores multi-source source-free adaptation for driving scene parsing, with an emphasis on reliable collaboration among heterogeneous source models for target-domain perception.

\subsection{Semi-Supervised Learning for Domain Adaptation}
\label{sec.related_works.ss_semantic}

Semi-supervised learning has been widely used in UDA to exploit unlabeled target-domain data for reducing domain gaps~\cite{RankMatch}.
Existing methods mainly follow consistency regularization~\cite{sshl,ssgan} or self-training~\cite{AllSpark,Saliency}.
Consistency regularization encourages prediction stability under data perturbations, but excessive or inappropriate augmentations may distort semantic structures in complex driving scenes~\cite{CorrMatch}.
Self-training methods iteratively refine semantic representations using pseudo labels as supervisory signals.
However, they typically rely on positive pseudo labels generated by a single model and overlook negative information that can help distinguish visually similar road-scene categories~\cite{Liang_2023_ICCV}.
Such single-model supervision can constrain adaptation, as the student model may inherit teacher-specific biases instead of benefiting from complementary knowledge across multiple source models~\cite{wang2024curriculum}.
To address these limitations, UCDA extends self-training into a collaborative multi-expert framework.
By converting diverse predictions from multiple source models into positive and negative consistency constraints, UCDA exploits both inter-model agreement and complementary negative cues, providing supervisory signals that are difficult to obtain from a single-model adaptation framework.

\subsection{Knowledge Distillation for Domain Adaptation}
\label{sec.related_works.kd}

Knowledge distillation has been widely used in domain adaptation to transfer knowledge from source models to a target model~\cite{Deng_2025_CVPR}.
Most existing methods rely on a single teacher model to guide adaptation.
They typically distill knowledge through logits~\cite{CoUDA}, pseudo labels~\cite{Lee_2025_CVPR}, or feature alignment~\cite{fdbu}, allowing the student model to inherit semantic structures learned from the source domain.
However, single-teacher distillation is inherently constrained by the capacity and bias of the teacher model~\cite{hiwk}.
Since the teacher is trained on a specific source dataset, its knowledge may reflect source-specific visual patterns and category distributions, which can be suboptimal for diverse target driving environments.
To mitigate this limitation, recent studies~\cite{yang2025multi,hiwk,Wen_2024_CVPR} have explored multi-teacher distillation to exploit complementary knowledge from multiple source models.
Nevertheless, existing multi-teacher frameworks still primarily rely on model agreement as positive supervision, while underusing informative discrepancies among teachers.
Such discrepancies may provide useful cues for distinguishing ambiguous or visually similar road-scene categories~\cite{yang2025multi}.
Moreover, many methods lack an explicit mechanism for modeling teacher reliability and inter-model relationships, leading to suboptimal integration of complementary expertise~\cite{Wen_2024_CVPR}.
When multiple teachers share common failure modes in challenging driving scenarios, their predictions may also become simultaneously unreliable.
These observations motivate our collaborative distillation strategy, which jointly exploits inter-model agreement and discrepancies to enable more reliable knowledge transfer for robust driving scene parsing.

\section{Methodology}
\label{sec.methodology}

\subsection{Preliminaries}
\label{sec.methodology.preliminary}
   
Given $K$ distinct source domains\footnote{The subscripts $s$ and $t$ denote the source and target domains, respectively.}, the $k$-th domain is associated with a labeled dataset $\mathcal{D}_{s}^{k}$, which contains ${N_s}$ pairs of RGB images $\boldsymbol{X}_{s}^k \in \mathbb{R}^{H \times W \times 3}$ and corresponding pixel-level semantic annotations $\boldsymbol{Y}_{s}^k \in \mathbb{R}^{H \times W }$ with respect to $C$ semantic categories. 
The target domain is associated with an unlabeled dataset $\mathcal{D}_{t}$ containing $N_t$ RGB images $\boldsymbol{X}_{t}\in \mathbb{R}^{H \times W \times 3}$ collected from the deployment environment, without pixel-level semantic annotations.
In contrast to source-dependent UDA methods that require access to labeled source data, the proposed framework performs adaptation using only the pre-trained source models $\{\mathcal{M}_{s}^{k}\}_{k=1}^{K}$, without relying on any original source samples.
The objective is to learn a target model $\mathcal{M}_{t}$ for driving scene parsing by distilling diverse and complementary expertise from multiple source models during adaptation, enabling the target model to perform independently in the target driving environment.

\begin{figure*}[t!]
    \centering
    \includegraphics[width=0.999\textwidth]{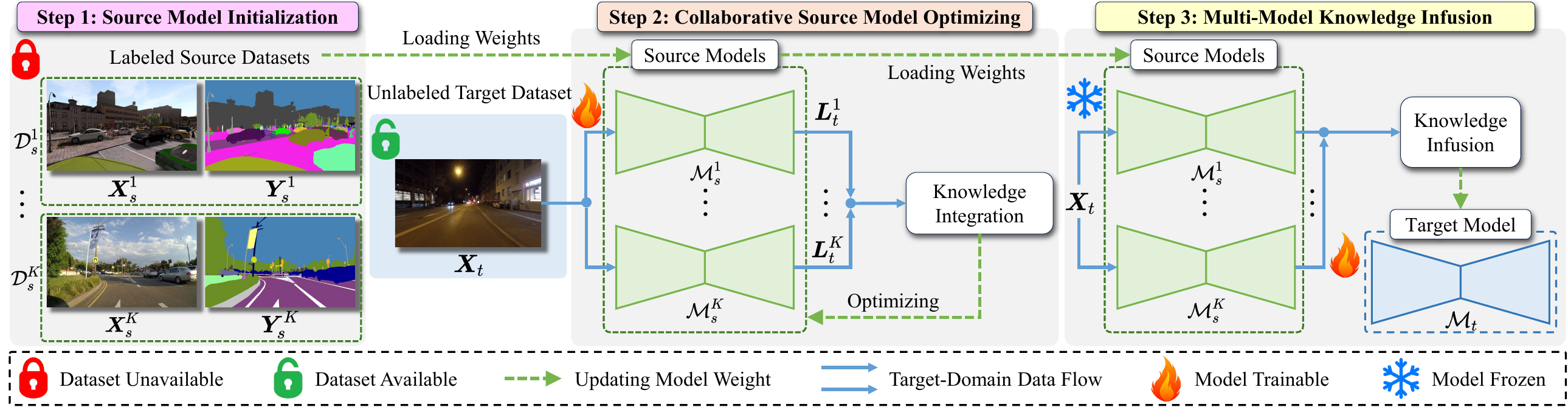}
    \caption{Overall architecture of the proposed unsupervised collaborative domain adaptation framework. Source model initialization in Step 1 follows a source-free adaptation setting where only pre-trained weights from diverse source domains are accessible. Using only unlabeled target-domain data, the framework learns a unified target model for reliable driving scene parsing.}
    \label{fig.framework}
\end{figure*}

\subsection{Framework Overview}
\label{sec.methodology.cca}

As illustrated in Fig.~\ref{fig.framework}, the proposed unsupervised collaborative domain adaptation framework leverages multiple pre-trained source models for source-free adaptation in driving scene parsing.
In Step 1, each source model is trained on its corresponding labeled source dataset, after which only the learned model parameters are retained and the original source data are discarded.
Adaptation is then conducted in two stages using unlabeled target-domain data.
In Step 2, collaborative source-model optimization refines the source models by integrating complementary knowledge across models.
In Step 3, multi-model knowledge distillation transfers the refined expertise into a unified target model for independent inference.
This design separates source-model refinement from final target-model learning, allowing complementary knowledge to be progressively validated and consolidated under the source-free constraint.
The following sections introduce the prototype-aware complementary knowledge synthesis module and the two-stage complementary knowledge transfer strategy.

\subsection{Prototype-Aware Complementary Knowledge Synthesis}
\label{sec.methodology.pgc}

Within the proposed UCDA framework, a central challenge is to establish a unified criterion for evaluating prediction reliability across independently trained source models, since their confidence estimates may be biased by source-specific data distributions and model-specific calibration behaviors.
To address this issue, the framework introduces a prototype-aware complementary knowledge synthesis mechanism, which uses class-level prototypes as shared semantic references to identify reliable predictions and derive complementary supervisory signals across source models.

\subsubsection{Prototype Memory Bank Initialization}
\label{sec.methodology.pgc.pmb}

To provide stable semantic references for cross-model reliability evaluation, a prototype memory bank is constructed by aggregating class-wise logit vectors from the predictions of each source model.
Specifically, for the $k$-th source model $\mathcal{M}_{s}^{k}$, the prototype vector $\boldsymbol{p}_{e}^k = \left( p_{e,1}^k, \dots, p_{e,C}^k \right)^\top\in\mathbb{R}^{C}$ is defined as the centroid of the logit vectors assigned to the $e$-th class with the following expression:
\begin{equation}
    \label{eq.initial_proto}
	p_{e,q}^k = \frac{\mathcal{S} \big({\boldsymbol{L}_{t,q}^{k} \odot \mathbbm{1}[\boldsymbol{Y}_{p}^{k} = e]\big)}   }{\mathcal{S} \big(\mathbbm{1}[\boldsymbol{Y}_{p}^{k} = e]\big)},
\end{equation}
where $\mathcal{S}(\cdot)$ denotes summation over all spatial locations, $\boldsymbol{L}_{t,q}^{k} \in \mathbb{R}^{H \times W}$ is the $q$-th channel of the target-domain logits $\boldsymbol{L}_{t}^{k} \in \mathbb{R}^{H \times W \times C}$ generated by $\mathcal{M}_s^k$, and $\boldsymbol{Y}_{p}^{k} \in \mathbb{R}^{H \times W}$ denotes the initial pseudo-label map obtained by applying the pixel-wise $\arg\max$ operation to the softmax prediction of $\mathcal{M}_{s}^{k}$ on target-domain images.
The indicator function $\mathbbm{1}[\boldsymbol{Y}_{p}^{k} = e]$ selects pixels assigned to the $e$-th class.
By concatenating all class prototypes, the prototype matrix of the $k$-th source model is obtained as $\boldsymbol{P}^k = \left( \boldsymbol{p}_{1}^k, \dots, \boldsymbol{p}_{C}^k \right)^\top \in \mathbb{R}^{C \times C}$.
The prototype matrices from all source models are then integrated to form the prototype memory bank.

\subsubsection{Reliability Quantification via Prototype Similarity}
\label{sec.methodology.pgc.cmce}

Pixel-wise reliability estimation across multiple source models is challenging because independently trained models often produce confidence scores on different scales due to source-specific biases.
To obtain a more consistent reliability criterion, the prototype memory bank provides class-level semantic anchors for comparing model predictions.
The reliability of each prediction is then quantified by measuring the similarity between its logit vector and the corresponding class prototype.
Specifically, for a pixel at spatial location $\boldsymbol{x}$, the reliability score $\boldsymbol{R}_{c}^k(\boldsymbol{x})$ of the $k$-th source model with respect to the $c$-th category is computed as:
\begin{equation}
\label{eq:cosine_sim}
\boldsymbol{R}_{c}^k(\boldsymbol{x}) =
\frac{ \left(\boldsymbol{L}_t^k(\boldsymbol{x})\right)^\top \boldsymbol{p}_{c}^k}
{\lVert\boldsymbol{L}_t^k(\boldsymbol{x})\rVert_2 \lVert\boldsymbol{p}_{c}^k\rVert_2},
\end{equation}
where $\boldsymbol{L}_t^k(\boldsymbol{x}) \in \mathbb{R}^{C}$ denotes the logit vector of the $k$-th source model at position $\boldsymbol{x}$, and $\boldsymbol{p}_{c}^k$ represents the corresponding prototype vector for the $c$-th category.
The resulting reliability map provides the basis for selecting trustworthy predictions and constructing complementary supervisory signals across source models.

\begin{figure*}[t!]
    \centering
    \includegraphics[width=0.999999\textwidth]{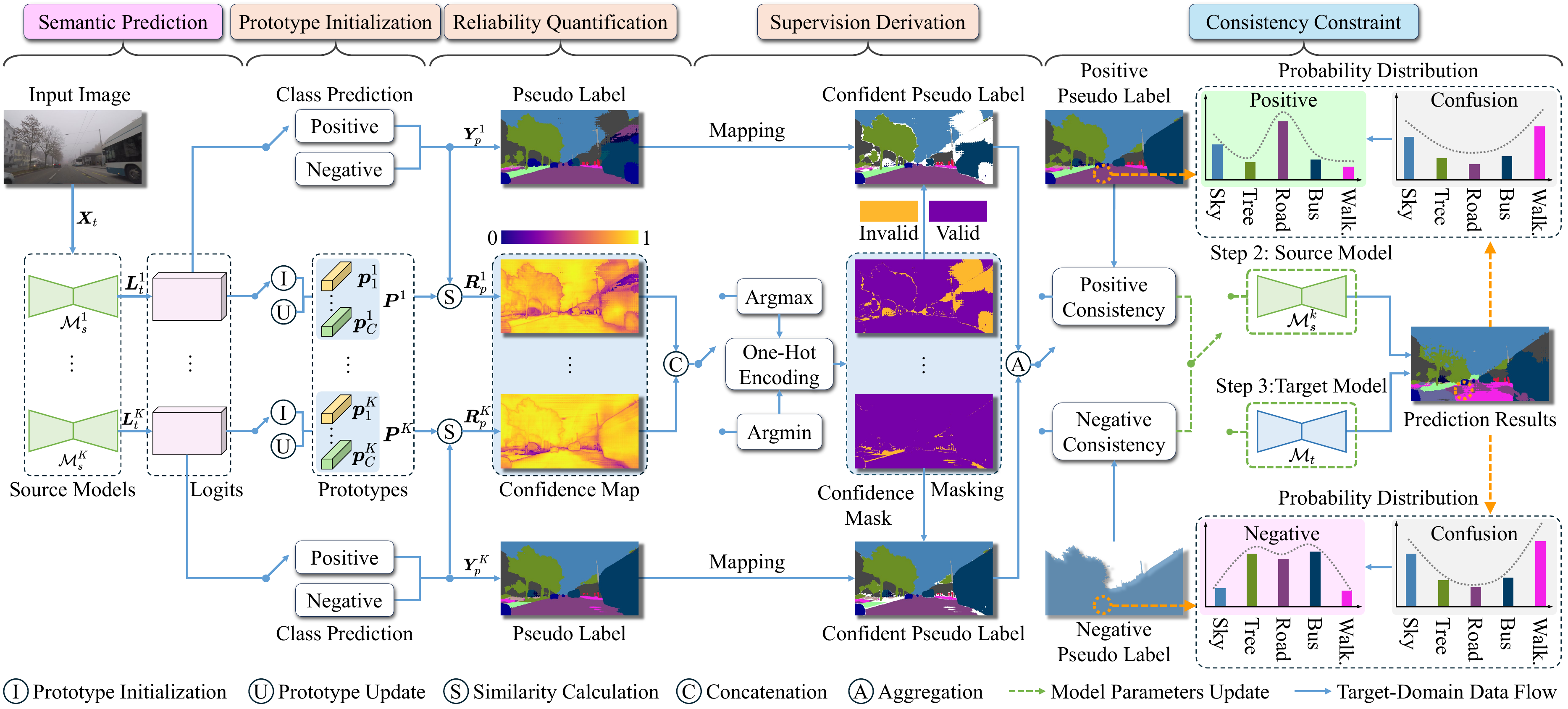}
    \caption{Detailed training pipeline of UCDA. Complementary supervisory signals are derived through cross-model reliability alignment and decoupled consistency constraints, supporting collaborative source-model refinement and final multi-source knowledge distillation into the target model.}
    \label{fig.de}
\end{figure*}

\subsubsection{Derivation of Complementary Supervision}
\label{sec.methodology.pgc.clpbu}

Based on the prototype-based reliability estimates, complementary supervision is constructed from multiple source models to support reliable cross-model optimization in the source-free adaptation setting.
For the $k$-th source model, the positive pseudo-label map $\boldsymbol{Y}_{p}^{k} \in \mathbb{R}^{H \times W}$ and the negative pseudo-label map $\boldsymbol{Y}_{n}^{k} \in \mathbb{R}^{H \times W}$ are obtained by selecting the categories with the maximum and minimum softmax probabilities, respectively.
Since these candidate labels can be affected by source-specific biases, a pixel-level selection process is required to identify trustworthy supervisory signals.
For each source model, $\boldsymbol{R}_p^k(\boldsymbol{x})$ and $\boldsymbol{R}_n^k(\boldsymbol{x})$ denote the reliability scores corresponding to its positive and negative pseudo labels at pixel $\boldsymbol{x}$, respectively.
Using the reliability scores computed in (\ref{eq:cosine_sim}), the source index $u$ is selected to determine the most reliable positive prediction, while $v$ is selected to identify the most reliable unlikely-category cue:
\begin{equation}
\label{eq.indices}
\begin{cases}
u = \underset{k\in \{1,\dots,K\} }{\arg\max} \ \boldsymbol{R}_{p}^{k}(\boldsymbol{x}), \\
v = \underset{k\in \{1,\dots,K\} }{\arg\min} \ \boldsymbol{R}_{n}^{k}(\boldsymbol{x}),
\end{cases}
\end{equation}
where $\boldsymbol{R}_{p,n}^{k} \in \mathbb{R}^{H \times W}$ denote the reliability maps associated with the positive pseudo-label map $\boldsymbol{Y}_{p}^{k}$ and the negative pseudo-label map $\boldsymbol{Y}_{n}^{k}$, respectively.

Based on this selection strategy, the discrete positive pseudo-label map $\boldsymbol{Y}_{p}$ is constructed by adopting the prediction from the most reliable source model $u$ at each spatial location $\boldsymbol{x}$, namely $\boldsymbol{Y}_{p}(\boldsymbol{x}) = \boldsymbol{Y}_{p}^{u}(\boldsymbol{x})$.
To preserve semantic structure in the predictive distributions, the corresponding logits are also extracted as soft supervisory signals for the selected positive and negative predictions, denoted as $\boldsymbol{L}_{p}(\boldsymbol{x}) = \boldsymbol{L}_{t}^{u}(\boldsymbol{x})$ and $\boldsymbol{L}_{n}(\boldsymbol{x}) = \boldsymbol{L}_{t}^{v}(\boldsymbol{x})$, respectively.
The discrete pseudo-label map provides pixel-level semantic supervision, while the soft supervisory signals retain class-wise distributional information for knowledge integration under inter-model discrepancies.
By integrating positive and negative information from multiple source models, the constructed supervision provides both semantic guidance and ambiguity suppression cues, forming the basis for subsequent cross-model consistency constraints.

\subsection{Two-Stage Complementary Knowledge Transfer Strategy}
\label{sec.methodology.cip}

Although complementary supervision from multiple source models provides useful semantic guidance, integrating it into a robust target model remains challenging.
Directly distilling unrefined source predictions may propagate source-specific biases and shared failure modes.
To address this issue, a two-stage transfer strategy is adopted, in which individual source models are first refined through collaborative optimization and then distilled into a unified target model.

\subsubsection{Source Model Initialization}
\label{sec.methodology.ccie}
In the source-free UDA setting, adaptation is performed using only pre-trained source model weights.
Before source-free adaptation, each source model is first trained on its corresponding labeled source domain to acquire domain-specific expertise.
Specifically, the $k$-th source model $\mathcal{M}_{s}^{k}$ is trained on $\mathcal{D}_{s}^{k}$ by minimizing the pixel-wise cross-entropy loss:
\begin{equation}
    \label{eq.cross_entropy_source}
	\mathcal{L}_{s}^k =  - \mathcal{S}\left({{\boldsymbol{H}}^{k}_{s}} \odot \log (\sigma(\boldsymbol{L}_{s}^{k}))\right),
\end{equation}
where ${\boldsymbol{H}}^{k}_{s} \in \{0,1\}^{H \times W \times C}$ denotes the one-hot encoding of $\boldsymbol{Y}_{s}^{k}$, $\odot$ is the Hadamard product, $\sigma(\cdot)$ denotes the softmax function, and $\boldsymbol{L}_{s}^k\in\mathbb{R}^{H\times W\times C}$ represents the logits produced by $\mathcal{M}_s^k$.

\subsubsection{Collaborative Source Model Optimization}
\label{sec.methodology.cmco}

The first stage refines each initialized source model using complementary supervision to mitigate domain-specific biases.
To this end, the supervision process is decoupled according to the confidence of the synthesized pseudo labels, enabling different optimization strategies for high-confidence and uncertain regions.
Specifically, spatial locations are partitioned into high- and low-confidence sets, denoted as $\mathcal{X}_{h}$ and $\mathcal{X}_{l}$, by applying a predefined threshold $\tau$ to the maximum softmax probability of the reliability-selected positive logits $\boldsymbol{L}_{p}$:
\begin{equation}
\begin{cases}
    \mathcal{X}_{h} = \{\boldsymbol{x} \mid \max \left(\sigma\left(\boldsymbol{L}_{p}(\boldsymbol{x})\right)\right) \ge \tau \}, \\
    \mathcal{X}_{l} = \{\boldsymbol{x} \mid \max \left(\sigma\left(\boldsymbol{L}_{p}(\boldsymbol{x})\right)\right) < \tau \}.
\end{cases}
\end{equation}
Binary masks $\boldsymbol{M}_{h,l} \in \{0, 1\}^{H \times W \times C}$ are then constructed to indicate high- and low-confidence regions, respectively, enabling cross-entropy supervision on reliable pixels and consistency regularization on ambiguous areas.
With these complementary supervision signals, each source model is optimized to reinforce reliable predictions while regularizing ambiguous regions.
For high-confidence regions, the synthesized pseudo-label map $\boldsymbol{Y}_{p}$ provides pixel-wise supervision through the following objective:
\begin{equation}
    \label{eq:ce_pseudo_refined}
	\mathcal{L}_{ce}^k =  - \mathcal{S}\left(\boldsymbol{M}_{h} \odot {{\boldsymbol{H}}_{t}} \odot \log (\sigma(\boldsymbol{L}_{t}^{k}))\right),
\end{equation}
where ${\boldsymbol{H}}_{t}$ denotes the one-hot encoding of the synthesized positive pseudo label $\boldsymbol{Y}_{p}$.

For ambiguous regions, discrete pseudo labels may be unreliable due to uncertain or conflicting predictions across source models. Therefore, dual consistency constraints are imposed to exploit soft supervisory signals and mitigate error accumulation.
In low-confidence regions, source-model predictions are aligned with the integrated positive soft targets through the following positive consistency loss:
\begin{equation}
\label{eq:positive_consistency_loss}
\mathcal{L}_{p}^k = - \mathcal{S}\left(
\boldsymbol{M}_{l} \odot
\sigma(\boldsymbol{L}_{p}) 
\odot \log (\sigma(\boldsymbol{L}_{t}^{k}))\right),
\end{equation}
where $\boldsymbol{M}_{l}$ denotes the mask for low-confidence regions where consistency regularization is applied.
This alignment preserves soft semantic cues from the predictive distributions, reducing the risk of enforcing incorrect hard labels when source models disagree.

In addition, a negative consistency loss is introduced in low-confidence regions to suppress unlikely class assignments indicated by $\boldsymbol{L}_{n}$:
\begin{equation}
\label{eq:negative_consistency_loss}
\begin{aligned}
\mathcal{L}_{n}^k = - \mathcal{S}\Big(
&\boldsymbol{M}_{l} \odot
(\mathbf{1}-{\sigma}(\boldsymbol{L}_{n})) \\
&\odot
\log (\mathbf{1}-{\sigma}(\boldsymbol{L}_{t}^{k}) + \epsilon)
\Big),
\end{aligned}
\end{equation}
where $\mathbf{1} \in \{1\}^{H \times W \times C}$ denotes an all-one tensor, and $\epsilon$ is a small constant for numerical stability.
This negative supervision complements the positive consistency constraint by discouraging implausible category assignments.
It explicitly suppresses categories that receive low probabilities from the selected negative supervisory signal.
Since each pixel has multiple negative categories but only one positive class, negative supervision provides additional constraints for resolving semantic ambiguity in uncertain regions.

Based on the above formulation, the complementary supervision signals are jointly optimized to refine each source model.
Specifically, the objectives for high- and low-confidence regions are integrated into the total loss for the $k$-th source model:
\begin{equation}
\label{eq:total_source_loss}
\mathcal{L}^k = \lambda_{ce}^k\mathcal{L}_{ce}^k + \lambda_{p}^k\mathcal{L}_{p}^k + \lambda_{n}^k\mathcal{L}_{n}^k,
\end{equation}
where $\lambda_{ce}^k$, $\lambda_{p}^k$, and $\lambda_{n}^k$ are hyperparameters that balance the loss components.
This collaborative refinement improves the reliability of each source model in the target domain while mitigating individual source biases, providing a more robust basis for subsequent knowledge distillation into the target model.

\subsubsection{Multi-Model Knowledge Infusion}
\label{sec.methodology.cki}

Collaborative refinement improves the reliability of the source models in the target domain, but their expertise remains distributed across separate models.
To obtain a unified target model for driving scene parsing, the refined source models are consolidated into a single model $\mathcal{M}_{t}$ for independent deployment in the target driving environment.
The refined source models are used as frozen teachers, and their enhanced expertise is transferred to the student model through uncertainty-aware distillation.
Specifically, the confidence-based region partition is guided by the refined positive logits $\boldsymbol{L}_{p}$.
These logits aggregate predictions from the optimized source models and provide a stable supervisory reference for student optimization.
In high-confidence regions, the student model is supervised by the synthesized pseudo labels through the cross-entropy loss $\mathcal{L}_{ce}$; in low-confidence regions, it is regularized by the positive and negative consistency losses $\mathcal{L}_{p}$ and $\mathcal{L}_{n}$.
The total training objective for the student model is formulated as:
\begin{equation}
\mathcal{L} = \lambda_{ce}\mathcal{L}_{ce} + \lambda_{p}\mathcal{L}_{p} + \lambda_{n}\mathcal{L}_{n},
\end{equation}
where $\mathcal{L}_{ce}$, $\mathcal{L}_{p}$, and $\mathcal{L}_{n}$ follow the formulations in (\ref{eq:ce_pseudo_refined}), (\ref{eq:positive_consistency_loss}), and (\ref{eq:negative_consistency_loss}), respectively, with the source-model logits $\boldsymbol{L}_{t}^{k}$ replaced by the student-model logits $\boldsymbol{L}_{t}^{\mathrm{stu}}$.
This distillation process yields a unified target model that integrates the collective expertise of the refined source models while reducing the influence of individual source biases.
As a result, the target model can perform robust scene parsing across diverse driving scenarios.

The proposed framework provides a unified strategy for adapting multiple source models under the source-free constraint, integrating complementary expertise while reducing error accumulation.
The effectiveness of this collaborative design for driving scene parsing is evaluated in the following section.

\section{Experiments}
\label{sec.exp}

\subsection{Datasets}
\label{sec.exp.dataset}

The evaluation covers cross-domain driving scene parsing tasks across synthetic-to-real and real-to-real adaptation, complemented by tests on data collected from an autonomous vehicle platform.
The selected datasets represent heterogeneous source environments and target deployment conditions, enabling assessment of UCDA under visual, semantic, and operational variations encountered by autonomous vehicles.
The source datasets include three synthetic datasets, GTA5~\cite{gta_dataset}, SYNTHIA~\cite{synthia_data}, and Synscapes~\cite{synscapes}, along with two large-scale real-world datasets, Mapillary Vistas~\cite{mapillary_data} and Cityscapes~\cite{cityscape}.
The synthetic datasets provide accurately annotated urban scenes that are difficult or costly to collect at scale in the real world, enabling synthetic-to-real evaluation.
Mapillary Vistas contains street-level scenes collected worldwide, covering varied weather, illumination, road layouts, and geographic conditions, which enrich the semantic and environmental diversity of the source models.
Cityscapes provides high-quality annotations of urban driving scenes and is widely used for driving scene parsing, supporting fair comparison with existing UDA methods.

The target domains include three real-world driving datasets, Cityscapes~\cite{cityscape}, BDD100K~\cite{bdd100}, and ACDC~\cite{acdc_data}, which represent deployment environments that autonomous vehicles may encounter after leaving the source training domains.
All target datasets are collected from real-world driving environments, ensuring that the evaluation reflects practical deployment scenarios and rigorously assesses the adaptability of the proposed framework to diverse and complex conditions.
Cityscapes provides finely annotated urban driving scenes and serves as a standard target environment for evaluating urban perception under domain shift, enabling fair comparison with existing UDA methods.
BDD100K contains large-scale driving images collected from multiple cities across North America, covering diverse weather conditions, times of day, and scene types, and therefore tests the generalization of UCDA across heterogeneous traffic environments.
ACDC focuses on adverse weather and challenging illumination conditions, including rain, snow, fog, and nighttime scenarios, providing a testbed for evaluating perception robustness under safety-critical and rarely observed operating conditions.
To support fair comparison and consistent assessment of dense perception across heterogeneous driving environments, all semantic categories are mapped to the 19- or 16-class label spaces defined by Cityscapes.

\subsection{Implementation Details}
\label{sec.exp.details}

The proposed framework is evaluated on both CNN-based and Transformer-based dense perception architectures. Specifically, DeepLab-V2~\cite{deeplabv2} with a ResNet-101 backbone~\cite{He_2016_CVPR} is used for conventional CNN-based evaluation, while MiT-B5 is adopted to assess the generality of UCDA on Transformer-based driving scene parsing models, following common settings in prior UDA studies~\cite{wang2024curriculum,Caoiapc}.
Following prior studies on UDA for driving scene parsing~\cite{wang2024curriculum,Caoiapc}, optimization is conducted using Stochastic Gradient Descent with a momentum of 0.9 and a weight decay of $5\times10^{-4}$.
The initial learning rates are set to $2.5\times10^{-4}$ for the backbone and $2.5\times10^{-3}$ for the classifier. Dense perception performance is evaluated using per-class Intersection-over-Union (IoU) and mean IoU (mIoU).
The symbol $^{\ast}$ indicates that challenging semantic categories are excluded, and mIoU$^{\ast}$ denotes the mean IoU over the remaining categories, providing a complementary evaluation protocol consistent with prior driving scene adaptation studies.
All experiments are implemented in PyTorch and conducted on five NVIDIA RTX 3090 GPUs.

\subsection{Comparison with State-of-the-Art Domain Adaptation Methods for Driving Scene Parsing}
\label{sec.exp.compare_SOTA}

\subsubsection{Adaptation Performance on the Cityscapes Dataset}
\label{sec.exp.compare_SOTA_city}

\begin{table*}[t!]
	\centering
	\setlength{\heavyrulewidth}{2pt}
    \setlength{\tabcolsep}{1.33mm} 
    \tablesmallfont
    \caption{Quantitative comparison of the proposed UCDA framework and SoTA UDA methods on the Cityscapes dataset (19 classes) based on per-class IoU ($\%$) and mIoU ($\%$). ``D'' and ``T'' denote adaptation using double and triple source models, respectively. Methods with the ``S-'' prefix denote source models used as baselines before adaptation.}
    \label{tab.cityscapes_19}
    \begin{tabular}{lc|c|ccccccccccccccccccc|c}
        \toprule[0.9pt]
        Method &Publication &\rotatebox[origin=c]{90}{Source-Free} & \rotatebox[origin=c]{90}{Road} & \rotatebox[origin=c]{90}{Sidewalk} & \rotatebox[origin=c]{90}{Building} & \rotatebox[origin=c]{90}{Wall} & \rotatebox[origin=c]{90}{Fence} & \rotatebox[origin=c]{90}{Pole} & \rotatebox[origin=c]{90}{Light} & \rotatebox[origin=c]{90}{Sign} & \rotatebox[origin=c]{90}{Vegetation} & \rotatebox[origin=c]{90}{Terrain} & \rotatebox[origin=c]{90}{Sky} & \rotatebox[origin=c]{90}{Person} & \rotatebox[origin=c]{90}{Rider} & \rotatebox[origin=c]{90}{Car} & \rotatebox[origin=c]{90}{Truck} & \rotatebox[origin=c]{90}{Bus} & \rotatebox[origin=c]{90}{Train} & \rotatebox[origin=c]{90}{Motorbike}& \rotatebox[origin=c]{90}{Bike} & \rotatebox[origin=c]{90}{mIoU} \\
        \midrule
        \multicolumn{23}{c}{\textbf{ResNet-101}} \\
        \midrule
        S-GTA5 &- &- &74.3	&26.2	&74.8	&21.6	&17.6	&20.0	&28.7	&16.9	&79.9	&24.1	&74.3	&57.1	&31.8	&65.0	&32.2	&29.7 &6.2	&30.7	&23.8	&38.7 \\
        
        S-SYNC &- &- &90.1	&48.3	&79.7	&30.8	&26.7	&31.1	&37.0	&38.7	&83.4	&37.8	&83.2	&56.4	&34.6	&84.5	&26.0	&27.7	&11.2	&32.1	&54.0	&48.1\\

        S-BDD &- &- &\textbf{94.9}	&\textbf{64.7}	&84.7	&37.1	&31.2	&33.6	&34.4	&40.1	&86.1	&\textbf{47.6}	&86.3	&62.0	&36.2	&88.2	&44.7	&50.1 &5.0	&33.3	&51.9	&53.3 \\

        \midrule
        CaCo~\cite{caco} &CVPR 2022 &\XSolidBrush &91.9 &54.3 &82.7 &31.7 &25.0 &38.1 &\textbf{46.7} &39.2 &82.6 &39.7 &76.2 &63.5 &23.6 &85.1 &38.6 &47.8 &10.3 &23.4 &35.1 &49.2 \\

        ARAS~\cite{aras} &T-CSVT 2023 &\XSolidBrush &91.9 &45.2 &81.8 &21.9 &25.6 &35.5 &41.5 &33.4 &85.1 &34.8 &73.8 &62.5 &31.6 &85.9 &33.8 &42.5 &7.3 &33.8 &42.8 &47.9 \\

        PBAL~\cite{pbal} &T-MM 2024 &\XSolidBrush &88.3 &38.2 &85.3 &32.5 &29.0 &37.5 &43.7 &40.5 &86.7 &33.5 &84.8 & \textbf{65.9} &28.6 &85.5 &33.3 &31.3 &26.8 &20.4 &48.6 &49.5 \\
        
        \midrule
        ATP~\cite{wang2024curriculum} &T-PAMI 2024 &\Checkmark &93.2 &55.8 &\textbf{86.5} &\textbf{45.2} &27.3 &36.6 &42.8 &37.9 &86.0 &43.1 &87.9 &63.5 &15.3 &85.5 &41.2 &\textbf{55.7} &0.0 &38.1 &57.4 &52.6 \\

        DBC~\cite{Tiandbc} &T-MM 2024 &\Checkmark &89.3 &38.7 &85.7 &34.1 &28.5 &\textbf{39.1} &43.5 &44.4 &86.2 &36.0 &84.5 &60.2 &25.2 &84.0 &38.2 &49.5 &6.5 &32.2 &45.9 &50.1\\

        IAPC~\cite{Caoiapc} &T-IV 2024  &\Checkmark &90.9 &36.5 &84.4 &36.1 &31.3 &32.9 &39.9 &38.7 &84.3 &38.6 &87.5 &58.6 &28.8 &84.3 &33.8 &49.5 &0.0 &34.1 &47.6 &49.4\\

        SND~\cite{SND} &CVPR 2024  &\Checkmark &93.0 &54.0 &84.6 &35.6 &30.3 &31.0 &41.9 &41.6 &\textbf{87.6} &44.6 &86.4 &62.6 &38.5 &87.5 &48.7 &42.9 &\textbf{36.6} &\textbf{49.5} &58.7 &55.6 \\

        \midrule
        \textbf{UCDA (D)} &- &\Checkmark &91.5	&53.0	&84.7	&35.4	&35.5	&26.0	&44.3	&41.9	&85.0	&39.7	&85.3	&63.0	&38.3	&85.5	&45.9	&47.6	&26.5	&45.1	&56.9	&54.3 \\

        \textbf{UCDA (T)} &- &\Checkmark &93.6	&60.7	&86.1	&43.2	&\textbf{38.9}	&30.6	&45.7	&\textbf{47.4}	&85.9	&42.5	&\textbf{88.4}	&64.3	&\textbf{40.8}	&\textbf{88.6}	&\textbf{57.1}	&52.2	&0.0	&44.4	&\textbf{58.8}	&\textbf{56.3} \\
        \midrule
        \multicolumn{23}{c}{\textbf{MiT-B5}} \\
        \midrule
        S-GTA5 &- &- &86.3 &29.3 &80.2 &31.8 &25.6 &24.0 &43.8 &19.1 &83.5 &36.0 &79.6 &64.0 &29.4 &84.6 &30.1 &23.1 &3.0 &29.3 &39.3 &44.3	\\
        S-SYNC &- &- &92.5 &49.9 &82.3 &33.0 &37.0 &46.9 &56.6 &57.5 &87.8 &36.9 &87.8 &67.8 &47.9 &88.7 &20.4 &28.9 &15.9 &29.7 &62.5 &54.2\\
        S-BDD &- &- &94.6 &66.3 &89.3 &25.0 &37.2 &50.9 &46.6 &55.1 &90.2 &44.0 &93.0 &68.1 &31.8 &92.7 &75.1 &64.8 &0.0 &42.6 &60.4 &59.4\\
        \midrule
        DAFormer~\cite{hoyer2022daformer} &CVPR 2022 &\XSolidBrush &95.7 &70.2 &89.4 &53.5 &48.1 &49.6 &55.8 &59.4 &89.9 &47.9 &92.5 &72.2 &44.7 &92.3 &74.5 &78.2 &65.1 &55.9 &61.8 &68.3 \\
        HRDA~\cite{hoyer2022hrda} &ECCV 2022 &\XSolidBrush &96.4 &74.4 &91.0 &\textbf{61.6} &51.5 &57.1 &\textbf{63.9} &\textbf{69.3} &91.3 &48.4 &\textbf{94.2} &\textbf{79.0} &\textbf{52.9} &93.9 &84.1 &\textbf{85.7} &75.9 &\textbf{63.9} &67.5 &\textbf{73.8}\\
        IDM~\cite{Wang_2023_ICCV} &ICCV 2023 &\XSolidBrush &\textbf{97.2} &\textbf{77.1} &89.8 &51.7 &\textbf{51.7} &54.5 &59.7 &64.7 &89.2 &45.3 &90.5 &74.2 &46.6 &92.3 &76.9 &59.6 &\textbf{81.2} &57.3 &62.4 &69.5 \\
        \midrule
        DAFormer~\cite{hoyer2022daformer} &CVPR 2022 &\Checkmark &87.7 &33.4 &83.9 &28.1 &27.5 &35.9 &42.9 &28.7 &82.4 &28.6 &83.1 &65.0 &37.0 &85.8 &53.9 &46.3 &31.8 &23.6 &36.8 &49.6 \\
        HRDA~\cite{hoyer2022hrda} &ECCV 2022 &\Checkmark &83.3 &28.2 &83.3 &43.3 &22.2 &42.9 &47.7 &38.2 &87.2 &40.0 &81.6 &69.5 &35.9 &84.8 &42.7 &50.4 &41.2 &33.7 &29.6 &51.9 \\
        IDM~\cite{Wang_2023_ICCV} &ICCV 2023 &\Checkmark &93.9 &59.1 &86.6 &35.3 &30.4 &42.2 &45.1 &57.8 &88.4 &35.1 &89.4 &69.7 &39.8 &89.1 &66.8 &46.0 &13.5 &41.1 &61.2 &57.4 \\
        ATP~\cite{wang2024curriculum} &T-PAMI 2024 &\Checkmark &96.6 &75.3 &89.4 &50.2 &41.5 &47.5 &48.6 &61.1 &89.8 &48.3 &93.4 &70.4 &40.1 &89.8 &66.7 &58.2 &30.3 &53.4 &65.6 &64.0\\
        \midrule
        \textbf{UCDA (D)} &- &\Checkmark &94.6 &66.3 &90.1 &37.6 &43.4 &55.3 &63.5 &64.5 &90.7 &53.8 &94.1 &73.4 &42.8 &93.4 &70.6 &69.0 &0.0 &54.9 &\textbf{67.8} &64.5 \\
        \textbf{UCDA (T)} &- &\Checkmark &95.1 &68.3 &\textbf{93.8} &43.1 &49.6 &\textbf{58.0} &61.8 &65.3 &\textbf{93.5} &\textbf{60.2} &94.1 &73.4 &42.9 &\textbf{95.8} &\textbf{84.5} &81.7 &3.1 &59.4 &67.6 &68.0 \\
        \bottomrule[0.9pt]
    \end{tabular}
\end{table*}

\begin{table*}[t!]
	\centering
	\setlength{\heavyrulewidth}{2pt}
    \setlength{\tabcolsep}{1.6mm}
    \tablesmallfont
    \caption{Quantitative comparison of the proposed UCDA framework and SoTA UDA methods on the Cityscapes dataset (16 classes) based on per-class IoU ($\%$) and mIoU ($\%$). ``D'' and ``T'' denote adaptation using double and triple source models, respectively. Methods with the ``S-'' prefix denote source models used as baselines before adaptation.}
    \label{tab.cityscapes_16}
    \begin{tabular}{lc|c|cccccccccccccccc|cc}
        \toprule[0.9pt]
        Method  &Publication &\rotatebox[origin=c]{90}{Source-Free}
        & \rotatebox[origin=c]{90}{Road} & \rotatebox[origin=c]{90}{Sidewalk} & \rotatebox[origin=c]{90}{Building} 
        & \rotatebox[origin=c]{90}{{Wall}$^{\ast}$} & \rotatebox[origin=c]{90}{Fence$^{\ast}$} & \rotatebox[origin=c]{90}{Pole$^{\ast}$} 
        & \rotatebox[origin=c]{90}{Light} & \rotatebox[origin=c]{90}{Sign} & \rotatebox[origin=c]{90}{Vegetation} 
        & \rotatebox[origin=c]{90}{Sky} & \rotatebox[origin=c]{90}{Person} & \rotatebox[origin=c]{90}{Rider} 
        & \rotatebox[origin=c]{90}{Car} & \rotatebox[origin=c]{90}{Bus} & \rotatebox[origin=c]{90}{Motorbike}
        & \rotatebox[origin=c]{90}{Bike} & \rotatebox[origin=c]{90}{mIoU} & \rotatebox[origin=c]{90}{mIoU$^{\ast}$} \\
        \midrule
        \multicolumn{21}{c}{\textbf{ResNet-101}} \\
        \midrule
        S-SYNS~\cite{deeplabv2} &T-PAMI 2018 &\XSolidBrush &78.1	&29.8	&75.0	&11.6	&0.2	&25.5	&10.5	&15.5	&71.1	&70.4	&55.8	&16.8	&64.2	&31.5	&17.2	&21.5	&37.2 &42.9\\
        S-GTA5~\cite{deeplabv2} &T-PAMI 2018 &\XSolidBrush &62.2	&24.4	&63.9	&20.8	&21.0	&26.1	&31.6	&20.4	&79.8	&74.5	&56.4	&31.4	&58.7	&42.3	&28.5	&20.9	&41.4 &45.8\\ 
        S-SYNC~\cite{deeplabv2} &T-PAMI 2018 &\XSolidBrush &89.4	&46.3	&79.8	&31.4	&25.5	&31.5	&34.5	&37.1	&84.1	&82.2	&58.8	&32.7	&83.9	&23.0	&29.6	&56.6	&51.6 &56.8 \\   
        \midrule
        CaCo~\cite{caco} &CVPR 2022  &\XSolidBrush &87.4 &48.9 &79.6 &8.8 &0.2 &30.1 &17.4 &28.3 &79.9 &81.2 &56.3 &24.2 &78.6 &39.2 &28.1 &48.3 &46.0 &53.6 \\

        ARAS~\cite{aras} &T-CSVT 2023 &\XSolidBrush &85.6 &39.2 &79.9 &15.5 &0.3 &32.2 &19.3 &23.9 &79.1 &81.7 &61.1 &19.3 &82.9 &25.7 &10.6 &51.9 &44.3 &50.8 \\
        
        PBAL~\cite{pbal} &T-MM 2024 &\XSolidBrush &85.3 &42.5 &81.7 &10.2 &0.1 &\textbf{36.8} &23.6 &31.8 &85.1 &87.6 &64.1 &27.7 &85.2 &31.4 &23.0 &35.6 &47.0 &54.2 \\
        
        \midrule
        ATP~\cite{wang2024curriculum} &T-PAMI 2024 &\Checkmark &\textbf{90.1} &46.3 &82.5 &0.0 &0.1 &31.7 &10.7 &17.9 &85.1 &\textbf{87.7} &\textbf{64.6} &\textbf{34.6} &86.4 &\textbf{54.8} &33.7 &\textbf{58.3} &49.0 &57.9 \\

        DBC~\cite{Tiandbc} &T-MM 2024 &\Checkmark &83.5 &45.8 &82.1 &7.8 &0.1 &30.9 &17.1 &24.1 &83.6 &86.4 &60.0 &20.2 &81.1 &37.2&9.2 &40.1 &44.3 &51.6 \\

        IAPC~\cite{Caoiapc} &T-IV 2024 &\Checkmark &68.5 &29.2 &82.0 &10.9 &1.2 &28.7 &22.3 &29.1 &82.8 &85.3 &60.5 &19.3 &83.1 &42.5 &32.2 &47.7 &45.3 &52.7 \\
        
        SND~\cite{SND} &CVPR 2024 &\Checkmark &88.1 &47.4 &80.1 &28.1 &32.2 &34.9 &33.6 &\textbf{41.3} &83.3 &86.7 &59.9 &27.2 &\textbf{86.7} &48.1 &36.2 &52.5 &54.1 &59.3 \\
        
        \midrule
        \textbf{UCDA (D)} &- &\Checkmark &\textbf{90.1}	&\textbf{50.8}	&83.7	&33.5	&26.6	&30.6	&42.1	&41.1	&\textbf{86.0}	&83.1	&60.3	&28.3	&84.6	&33.3	&\textbf{41.9}	&57.5	&54.6 &60.2\\
        \textbf{UCDA (T)} &- &\Checkmark &89.5	&48.4	&\textbf{85.4}	&\textbf{34.6}	&\textbf{34.6}	&29.3	&\textbf{44.0}	&35.7	&\textbf{86.0}	&87.1	&63.5	&33.2	&85.2	&50.4	&41.3	&56.2	&\textbf{56.5} &\textbf{62.0}\\
        \midrule
        \multicolumn{21}{c}{\textbf{MiT-B5}} \\
        \midrule
        S-SYNS~\cite{deeplabv2} &T-PAMI 2018 &\XSolidBrush &81.9 &38.6 &83.6 &22.3 &0.4 &44.7 &39.7 &29.7 &86.4 &78.2 &65.1 &25.9 &82.4 &46.4 &30.2 &34.2 &49.3 &55.5	\\
        S-GTA5~\cite{deeplabv2} &T-PAMI 2018 &\XSolidBrush &80.7 &31.3 &79.4 &29.7 &24.8 &34.3 &43.6 &19.9 &83.1 &79.7 &61.1 &31.5 &79.9 &44.2 &32.6 &37.4 &49.6 &54.2\\
        S-SYNC~\cite{deeplabv2} &T-PAMI 2018 &\XSolidBrush &92.8 &53.0 &81.7 &32.6 &34.0 &45.4 &52.1 &56.9 &87.3 &87.6 &66.4 &43.9 &91.0 &41.1 &30.5 &63.0 &60.0 &65.2\\
        \midrule
        DAFormer~\cite{hoyer2022daformer} &CVPR 2022 &\XSolidBrush &84.5 &40.7 &88.4 &41.5 &6.5 &50.0 &55.0 &54.6 &86.0 &89.8 &73.2 &48.2 &87.2 &53.2 &53.9 &61.7 &60.9 &67.4 \\
        HRDA~\cite{hoyer2022hrda} &ECCV 2022 &\XSolidBrush &85.2 &47.7 &\textbf{88.8} &\textbf{49.5} &4.8 &\textbf{57.2} &\textbf{65.7} &60.9 &85.3 &92.9 &\textbf{79.4} &\textbf{52.8} &89.0 &64.7 &\textbf{63.9} &64.9 &65.8 &72.4\\
        IDM~\cite{Wang_2023_ICCV} &ICCV 2023 &\XSolidBrush &87.6 &47.6 &88.1 &33.4 &6.3 &52.8 &57.8 &56.5 &83.0 &77.5 &66.2 &52.1 &89.3 &55.6 &57.1 &64.2 &60.9 &67.9 \\
        \midrule
        DAFormer~\cite{hoyer2022daformer} &CVPR 2022 &\Checkmark &64.3 &25.1 &78.5 &23.8 &1.9 &37.3 &29.7 &22.8 &80.4 &83.0 &65.1 &26.6 &69.8 &38.3 &22.7 &32.8 &43.8 &49.2 \\
        HRDA~\cite{hoyer2022hrda} &ECCV 2022 &\Checkmark &72.2 &26.6 &80.8 &23.0 &0.5 &42.5 &41.0 &31.5 &84.3 &86.2 &64.3 &29.3 &73.5 &28.8 &12.4 &41.6 &46.1 &51.3 \\
        IDM~\cite{Wang_2023_ICCV} &ICCV 2023 &\Checkmark &82.2 &37.9 &83.5 &20.3 &1.5 &47.3 &41.7 &25.6 &84.4 &86.8 &61.6 &25.0 &87.6 &43.7 &30.2 &36.4 &49.7 &55.9 \\
        ATP~\cite{wang2024curriculum} &T-PAMI 2024 &\Checkmark &90.6 &54.4 &86.7 &28.5 &0.5 &50.3 &52.4 &50.5 &87.4 &\textbf{93.4} &70.2 &35.8 &89.6 &53.5 &50.6 &51.1 &59.1 &66.6\\
        \midrule
        \textbf{UCDA (D)} &- &\Checkmark &\textbf{93.6} &\textbf{60.4} &88.3 &30.5 &35.2 &54.2 &62.1 &\textbf{62.0} &90.6 &91.8 &72.9 &46.1 &\textbf{92.0} &64.2 &54.3 &\textbf{69.4} &66.7 &\textbf{72.9} \\
        \textbf{UCDA (T)} &- &\Checkmark &92.9 &58.9 &88.3 &46.8 &\textbf{38.3} &52.0 &60.3 &56.5 &\textbf{90.7} &91.7 &73.8 &46.1 &\textbf{92.0} &\textbf{73.4} &53.4 &61.3 &\textbf{67.3} &72.2\\
    \bottomrule[0.9pt]
    \end{tabular}
\end{table*}

\begin{figure*}[t!]
    \centering
    \includegraphics[width=0.999999\textwidth]{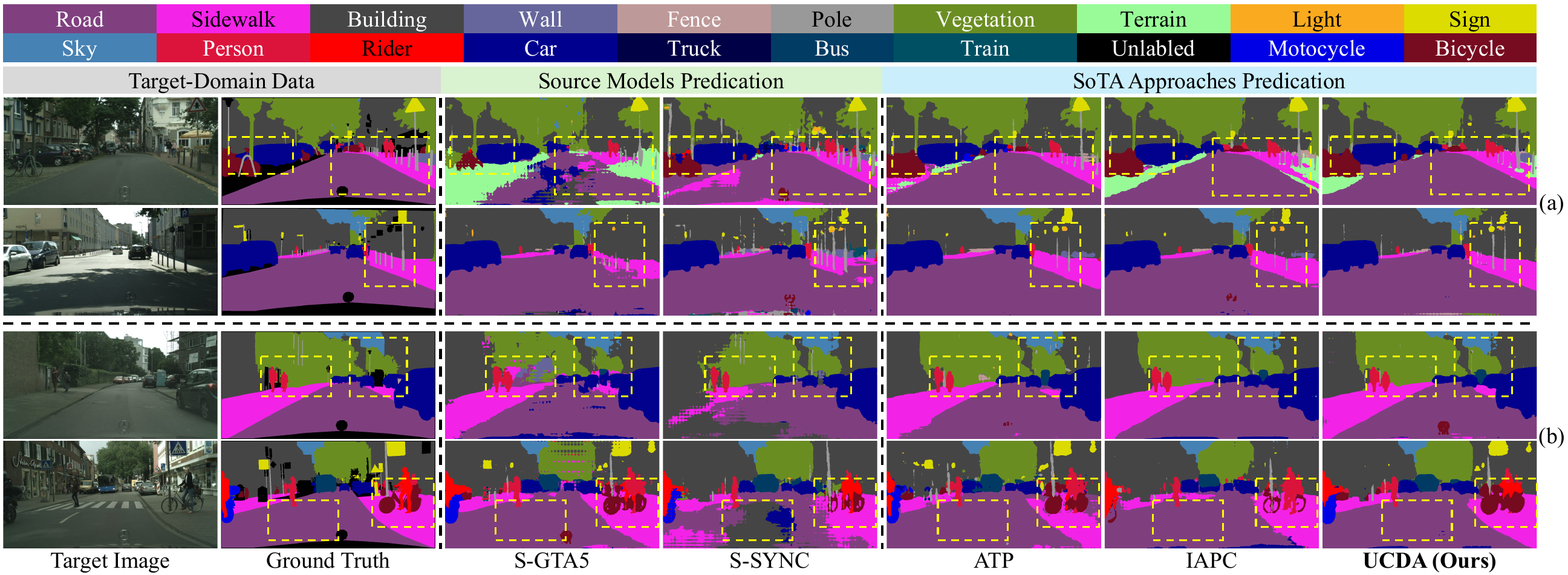}
    \vspace{-18pt} 
    \caption{Qualitative results for (a) 19-class and (b) 16-class parsing tasks on the Cityscapes dataset.}
    \label{fig.city19}
\end{figure*}

Tables~\ref{tab.cityscapes_19} and~\ref{tab.cityscapes_16} report the quantitative results on the Cityscapes~\cite{cityscape} dataset under the 19-class and 16-class evaluation protocols, respectively, while Fig.~\ref{fig.city19} presents the corresponding qualitative results. The two protocols support comparison with different groups of existing domain adaptation methods and jointly evaluate the ability of UCDA to transfer dense perception knowledge to an urban target environment without access to source-domain data.

As shown in Tables~\ref{tab.cityscapes_19} and~\ref{tab.cityscapes_16}, UCDA improves target-domain parsing over source-only models on both CNN-based and Transformer-based architectures, especially for categories that are often underrepresented or sensitive to domain shift, such as fence, traffic light, truck, and motorbike.
These gains indicate that complementary source models provide useful category-specific knowledge for urban driving perception when the original source data are unavailable.
Under the ResNet-101 setting, UCDA achieves the best overall mIoU in both the 19-class and 16-class protocols, outperforming existing source-free methods and also surpassing the compared source-dependent methods despite not using source-domain data during adaptation.
Under the MiT-B5 setting, UCDA obtains the best performance among source-free methods, while remaining competitive with source-dependent methods that still rely on access to labeled source data.
Although certain dominant classes may show marginal degradation compared with the strongest individual model or source-dependent counterpart, detailed analysis is provided in Sect.~\ref{sec.discussion}.
This behavior reflects the remaining difficulty of preserving all source-domain strengths during source-free collaboration, but the overall results still demonstrate that UCDA effectively consolidates complementary perception knowledge across heterogeneous source models for reliable dense perception in the target driving domain.
Compared with existing source-free methods~\cite{wang2024curriculum,Tiandbc,Caoiapc,SND}, UCDA achieves the strongest overall results under the ResNet-101 setting and maintains clear advantages under the MiT-B5 setting, indicating that the proposed collaborative strategy is effective across both CNN-based and Transformer-based dense perception architectures.
For several difficult categories, such as wall and fence, some existing source-free methods produce very low IoU scores, revealing their limited ability to recover target-domain semantics for visually ambiguous or underrepresented road-scene structures.
This limitation is mainly caused by the insufficient category-specific knowledge retained in a single source model, especially when the target driving environment contains scene layouts or object appearances that are weakly represented in the corresponding source domain.
In contrast, UCDA exploits complementary predictions from multiple source models, allowing reliable category cues from one source model to compensate for the weak or uncertain predictions of another.
This collaborative compensation mechanism is a key advantage of UCDA in source-data-unavailable driving perception scenarios.
The results also show that increasing the number of source models generally improves target-domain performance, suggesting that more diverse source expertise provides richer supervision for adaptation.
Specifically, UCDA with triple source models usually outperforms the double-source variant, especially in overall mIoU and in categories that benefit from complementary source-domain knowledge.
The influence of source-model diversity and source-model quantity is further analyzed in Sect.~\ref{sec.discussion}.

\subsubsection{Adaptation Performance on the BDD100K Dataset}
\label{sec.exp.compare_SOTA_bdd}

\begin{table*}[t!]
	\centering
	\setlength{\heavyrulewidth}{2pt}
    \setlength{\tabcolsep}{1.85mm} 
    \tablesmallfont
    \caption{Quantitative comparison of the proposed UCDA framework and SoTA UDA methods on the BDD100K dataset (16 classes) based on per-class IoU ($\%$) and mIoU ($\%$). ``D'' and ``T'' denote adaptation using double and triple source models, respectively. Methods with the ``S-'' prefix denote source models used as baselines before adaptation.}
    \label{tab.bdd_16}
    \begin{tabular}{lc|c|cccccccccccccccc|c}
        \toprule[0.9pt]
        Method &Publication &\rotatebox[origin=c]{90}{Source-Free}
        & \rotatebox[origin=c]{90}{Road} & \rotatebox[origin=c]{90}{Sidewalk} & \rotatebox[origin=c]{90}{Building} 
        & \rotatebox[origin=c]{90}{Wall} & \rotatebox[origin=c]{90}{Fence} & \rotatebox[origin=c]{90}{Pole} 
        & \rotatebox[origin=c]{90}{Light} & \rotatebox[origin=c]{90}{Sign} & \rotatebox[origin=c]{90}{Vegetation} 
        & \rotatebox[origin=c]{90}{Sky} & \rotatebox[origin=c]{90}{Person} & \rotatebox[origin=c]{90}{Rider} 
        & \rotatebox[origin=c]{90}{Car} & \rotatebox[origin=c]{90}{Bus} & \rotatebox[origin=c]{90}{Motorbike}
        & \rotatebox[origin=c]{90}{Bike} & \rotatebox[origin=c]{90}{mIoU} \\
        \midrule
        \multicolumn{20}{c}{\textbf{ResNet-101}} \\
        \midrule
        S-GTA5~\cite{deeplabv2} &T-PAMI 2018 &\XSolidBrush &50.8	&27.2	&48.5 &5.4	&27.9	&27.7	&30.7	&18.7	&64.8	&66.4	&53.3	&32.8	&65.4	&20.2	&35.4	&20.9	&37.3 \\

        S-SYNC~\cite{deeplabv2} &T-PAMI 2018 &\XSolidBrush &79.1	&32.3	&65.8 &3.8 &8.5	&28.5	&33.9	&26.4	&76.7	&86.9	&33.6	&22.1	&79.1	&15.2	&15.9	&30.7	&39.9 \\

        S-CITY~\cite{deeplabv2} &T-PAMI 2018 &\XSolidBrush &86.3	&\textbf{52.8}	&63.1	&14.2	&27.7	&32.8	&37.3	&\textbf{42.4}	&81.1	&73.5	&\textbf{57.1}	&\textbf{37.3}	& \textbf{86.1}	&39.9	&\textbf{47.1}	&\textbf{46.5}	&51.6 \\
        \midrule
        URMDA~\cite{urma} &CVPR 2021 &\Checkmark &83.9 &38.3 &78.7 &9.6 &7.3 &29.1 &11.1 &4.9 &70.7 &74.2 &53.8 &15.0 &81.2 &35.0 &22.8 &30.5 &40.4 \\
        
        HCL~\cite{hcl} &NeurIPS 2021 &\Checkmark &\textbf{88.6} &39.2 &\textbf{81.0} &8.2 &7.9 &28.4 &11.4 &5.7 &71.0 &77.2 &54.2 &16.0 &81.8 &41.4 &22.6 &31.4 &41.6 \\

        SFDASEG~\cite{SFDASEG} &ICCV 2021 &\Checkmark &87.9 &40.2 &80.6 &13.1 &8.2 &30.2 &22.8 &17.1 &71.1  &78.1 &51.4 &27.9 &80.2  &43.7  &30.3 &42.3 &45.3 \\
        
        DTST~\cite{DTST} &CVPR 2023 &\Checkmark &83.1 &39.9 &64.9 &8.9 &14.5 &29.5 &27.0 &27.1 &71.9 &83.2 &52.9 &31.3 &74.7 &41.1 &30.3 &42.1 &45.2 \\
        
        SND~\cite{SND} &CVPR 2024 &\Checkmark &84.1 &42.6 &74.1 &15.2 &21.2 &31.1 &31.0 &25.5 &70.4 &83.9 &52.8 &33.9 &79.9 &39.1 &37.5 &41.9 &47.8 \\
        \midrule
        \textbf{UCDA (D)} &- &\Checkmark &77.4	&38.7	&66.1	&7.1	&31.8	&30.2	&\textbf{41.2}	&35.2	&78.0	&88.2	&56.2	&28.5	&83.4	&34.5	&37.8	&44.5	&48.7 \\
        \textbf{UCDA (T)} &- &\Checkmark &86.1	&51.4	&71.7	&\textbf{16.4}	&\textbf{38.1}	&\textbf{33.3}	&41.1	&36.4	&\textbf{82.6}	&\textbf{90.2}	&55.4	&27.4	&\textbf{86.1}	&\textbf{50.2}	&34.7	&45.1	&\textbf{52.9} \\
        \bottomrule[0.9pt]
    \end{tabular}
\end{table*}

\begin{figure*}[t!]
    \centering
    \includegraphics[width=0.999999\textwidth]{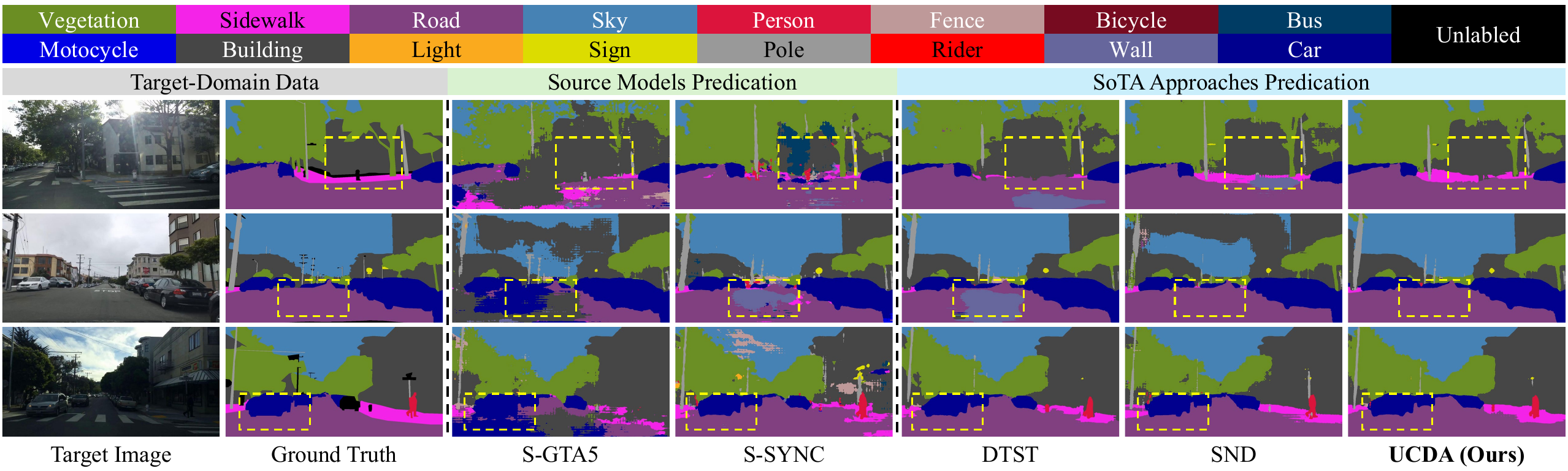}
    \vspace{-18pt} 
    \caption{Qualitative results on the BDD100K dataset across 16 semantic categories.}
    \label{fig.bdd16}
\end{figure*}

Table~\ref{tab.bdd_16} reports the quantitative results on the BDD100K~\cite{bdd100} dataset under the 16-class evaluation protocol, while Fig.~\ref{fig.bdd16} presents the corresponding qualitative results.
The results further evaluate the ability of UCDA to adapt dense perception models to a heterogeneous real-world driving target domain.
Compared with existing source-free methods, UCDA shows a larger performance gain on BDD100K than on Cityscapes, indicating stronger benefits under more diverse target-domain conditions.
BDD100K contains broader variations in city layout, weather, illumination, and traffic scenes, which impose higher generalization demands on driving scene parsing models and expose the limitations of relying on a single source model.
By integrating complementary expertise from multiple source models, UCDA mitigates source-specific biases and improves target-domain perception under complex and heterogeneous driving conditions.
The source-only results also show that S-SYNC and S-GTA5 perform worse than S-CITY on BDD100K, reflecting the larger domain gap between synthetic source domains and the real-world BDD100K target domain.
With three source models, UCDA surpasses the stronger S-CITY baseline, indicating that the weaker synthetic-source models still provide useful complementary cues when their predictions are integrated through reliability-aware collaboration.
This behavior allows UCDA to preserve performance on categories where the strongest source model is reliable while improving categories where individual source models are weak, leading to more balanced target-domain perception without requiring target annotations.

\subsubsection{Adaptation Performance on the ACDC Dataset}
\label{sec.exp.compare_SOTA_acdc}

\begin{table*}[t]
	\centering
	\setlength{\heavyrulewidth}{2pt}
    \setlength{\tabcolsep}{1.4mm} 
    \tablesmallfont
    \caption{Quantitative comparison of the proposed UCDA framework and SoTA UDA methods on the ACDC dataset (19 classes) based on per-class IoU ($\%$) and mIoU ($\%$). ``D'' and ``T'' denote adaptation using double and triple source models, respectively. Methods with the ``S-'' prefix denote source models used as baselines before adaptation.}
    \label{tab.acdc_19}
    \begin{tabular}{lc|c|ccccccccccccccccccc|c}
        \toprule[0.9pt]
        Method &Publication &\rotatebox[origin=c]{90}{Source-Free} & \rotatebox[origin=c]{90}{Road} & \rotatebox[origin=c]{90}{Sidewalk} & \rotatebox[origin=c]{90}{Building} & \rotatebox[origin=c]{90}{Wall} & \rotatebox[origin=c]{90}{Fence} & \rotatebox[origin=c]{90}{Pole} & \rotatebox[origin=c]{90}{Light} & \rotatebox[origin=c]{90}{Sign} & \rotatebox[origin=c]{90}{Vegetation} & \rotatebox[origin=c]{90}{Terrain} & \rotatebox[origin=c]{90}{Sky} & \rotatebox[origin=c]{90}{Person} & \rotatebox[origin=c]{90}{Rider} & \rotatebox[origin=c]{90}{Car} & \rotatebox[origin=c]{90}{Truck} & \rotatebox[origin=c]{90}{Bus} & \rotatebox[origin=c]{90}{Train} & \rotatebox[origin=c]{90}{Motorbike}& \rotatebox[origin=c]{90}{Bike} & \rotatebox[origin=c]{90}{mIoU} \\
        \midrule
        \multicolumn{23}{c}{\textbf{ResNet-101}} \\
        \midrule
        S-CITY~\cite{deeplabv2} &T-PAMI 2018 &\XSolidBrush &79.7	&42.4	&57.5	&26.3	&25.1	&30.6	&39.8	&34.8	&73.5	&25.5	&68.2	&41.5	&8.7	&68.6	&48.0	&29.7	&52.3	&24.4	&26.1	&42.3 \\

        S-BDD~\cite{deeplabv2} &T-PAMI 2018 &\XSolidBrush &84.0	&47.0	&70.7	&21.3	&25.2	&34.4	&36.0	&29.7	&76.6	&26.2	&89.4	&37.7	&8.4	&75.5	&59.5	&33.3	&3.9	&19.2	&15.2	&41.7 \\

        S-MAP~\cite{deeplabv2} &T-PAMI 2018 &\XSolidBrush &84.8	&48.6	&75.5	&33.5	&33.6	&40.3	&44.8	&35.3	&78.4	&27.1	&91.0	&44.2 &9.5	&\textbf{78.5}	&62.6	&51.2	&44.7	&26.8	&16.8	&48.8 \\

        \midrule
        DACS~\cite{dacs} &WACV 2021 &\XSolidBrush &58.5 &34.7 &76.4 &20.9 &22.6 &31.7 &32.7 &46.8 &58.7 &39.0 &36.3 &43.7 &20.5 &72.3 &39.6 &34.8 &51.1 &24.6 &38.2 &41.2 \\

        VBLC~\cite{vblc} &AAAI 2023 &\XSolidBrush &49.6 &39.3 &\textbf{79.4} &35.8 &29.5 &42.6 &\textbf{57.2} &57.5 &{69.1} &42.7 &39.8 &\textbf{54.5} &29.3 &77.8 &43.0 &36.2 &32.7 &38.7 &\textbf{53.4} &47.8 \\
        
        \midrule
        URMA~\cite{urma} &CVPR 2021 &\Checkmark &85.4 &52.9 &62.9 &20.4 &\textbf{34.4} &39.9 &36.7 &43.9 &74.9 &\textbf{46.9} &85.1 &27.2 &22.4 &76.0 &40.5 &41.5 &38.9 &20.6 &46.2 &47.2 \\
        
        HCL~\cite{hcl} &NeurIPS 2021 &\Checkmark &80.5 &42.9 &57.6 &14.7 &29.4 &40.3 &49.0 &51.1 &72.4 &35.6 &78.3 &39.7 &\textbf{31.8} &76.0 &35.4 &42.7 &42.5 &25.7 &43.0 &46.8 \\
        
        SimT~\cite{simt} &CVPR 2022 &\Checkmark &83.5 &52.7 &60.7 &19.6 &33.7 &42.0 &43.1 &47.4 &75.0 &42.5 &85.8 &39.8 &19.6 &76.9 &39.6 &42.7 &41.1 &24.0 &43.1 &48.0 \\
        
        ATP~\cite{wang2024curriculum} &T-PAMI 2024 &\Checkmark &76.2 &47.3 &71.4 &\textbf{42.7} &31.4 &\textbf{44.2} &55.4 &\textbf{62.0} &\textbf{89.0} &34.7 &79.1 &49.9 &16.6 &77.5 &30.0 &19.7 &47.7 &\textbf{44.0} &39.4 &50.5 \\
        \midrule
        \textbf{UCDA (D)} &- &\Checkmark &87.8	&\textbf{59.1}	&71.4	&37.5	&33.0	&39.8	&41.7	&37.6	&78.7	&33.0	&88.7	&43.4	&9.0	&76.9	&57.1	&\textbf{68.8}	&\textbf{70.5}	&22.7	&23.3	&51.6 \\
        \textbf{UCDA (T)} &- &\Checkmark &\textbf{88.3}	&58.1	&79.2	&38.8	&34.1	&38.5	&44.2	&37.9	&80.2	&29.8	&\textbf{92.1}	&43.7	&8.0	&78.4	&\textbf{64.3}	&68.0	&68.9	&26.7	&24.9	&\textbf{52.8} \\
        \midrule
        \multicolumn{23}{c}{\textbf{MiT-B5}} \\
        \midrule
        S-CITY~\cite{deeplabv2} &T-PAMI 2018 &\XSolidBrush &79.9 &42.4 &72.5 &36.9 &27.1 &48.8 &65.6 &55.5 &80.6 &38.9 &84.7 &56.7 &26.6 &80.0 &62.8 &70.2 &54.5 &44.9 &43.8 &56.4 \\
        S-BDD~\cite{deeplabv2} &T-PAMI 2018 &\XSolidBrush &83.2 &40.6 &79.9 &29.4 &31.9 &51.8 &58.9 &46.6 &81.4 &33.7 &91.0 &56.6 &33.2 &83.8 &63.7 &53.1 &12.2 &24.1 &23.6 &51.5\\
        S-MAP~\cite{deeplabv2} &T-PAMI 2018 &\XSolidBrush  &87.6 &58.0 &80.0 &44.3 &39.0 &54.0 &69.7 &53.4 &81.9 &27.6 &91.1 &56.4 &20.4 &85.7 &78.8 &80.1 &42.4 &32.3 &28.9 &58.5\\
        \midrule
        TENT~\cite{wang2021tent} &ICLR 2021 &\Checkmark &85.3 &50.2 &85.4 &45.4 &32.7 &50.4 &59.4 &66.1 &86.4 &45.7 &97.5 &57.9 &53.8 &84.7 &51.0 &66.9 &72.4 &40.2 &50.1 &62.2\\
        CoTTA~\cite{Wang_2022_CVPR} &CVPR 2022 &\Checkmark &85.7 &50.9 &85.9 &45.9 &33.6 &54.8 &62.3 &\textbf{69.9} &87.1 &45.7 &97.7 &63.3 &\textbf{59.4} &85.1 &52.8 &68.0 &74.1 &44.9 &55.1 &64.4\\
        IDM~\cite{Wang_2023_ICCV} &ICCV 2023 &\Checkmark &88.8 &63.2 &85.8 &45.5 &30.3 &42.1 &69.7 &62.7 &87.4 &\textbf{51.7} &96.5 &61.8 &29.3 &86.5 &\textbf{80.5} &68.9 &70.1 &\textbf{57.0} &52.8 &64.8 \\
        ATP~\cite{wang2024curriculum} &T-PAMI 2024 &\Checkmark &88.4 &63.2 &87.4 &50.0 &32.5 &53.5 &71.8 &66.9 &90.3 &42.7 &97.0 &\textbf{66.3} &32.1 &85.8 &63.9 &70.5 &\textbf{83.2} &46.1 &53.8 &65.6\\
        \midrule
        \textbf{UCDA (D)} &- &\Checkmark &89.6 &63.0 &88.4 &52.8 &\textbf{43.2} &\textbf{58.9} &\textbf{73.2} &64.5 &\textbf{90.5} &47.1 &98.1 &64.5 &27.5 &87.0 &68.2 &78.3 &\textbf{83.2} &43.7 &54.3 &67.2\\
        \textbf{UCDA (T)} &- &\Checkmark &\textbf{90.9} &\textbf{66.8} &\textbf{88.7} &\textbf{53.1} &43.0 &58.0 &71.5 &66.1 &\textbf{90.5} &48.6 &\textbf{98.2} &65.6 &37.0 &\textbf{89.5} &76.2 &\textbf{81.7} &78.9 &55.5 &\textbf{57.5} &\textbf{69.3}\\
        \bottomrule[0.9pt]
    \end{tabular}
\end{table*}

\begin{figure*}[t]
    \centering
    \includegraphics[width=0.999999\textwidth]{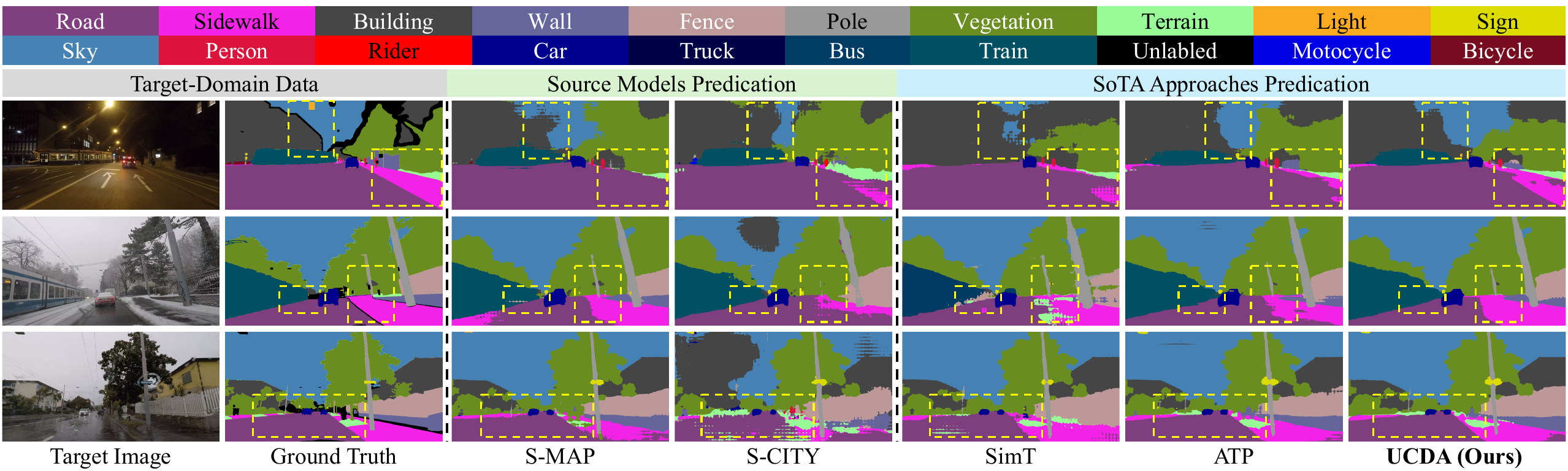}
    \vspace{-18pt} 
    \caption{Qualitative results on the ACDC dataset across 19 semantic categories.}
    \label{fig.acdc19}
\end{figure*}

Table~\ref{tab.acdc_19} reports the quantitative results on the ACDC dataset~\cite{acdc_data} under the 19-class evaluation protocol, while Fig.~\ref{fig.acdc19} presents the corresponding qualitative results.
The proposed UCDA achieves the highest overall mIoU under both ResNet-101 and MiT-B5 settings, demonstrating its ability to consolidate multi-source perception knowledge for reliable dense scene understanding under adverse driving conditions.
Under the ResNet-101 setting, UCDA achieves the best overall mIoU among all compared methods and improves several safety-relevant categories, including road, sidewalk, bus, and train.
Compared with existing source-free methods and adverse-condition adaptation methods~\cite{dacs,hcl,urma}, UCDA shows stronger robustness by integrating complementary source knowledge rather than relying on a single source model.
Even when individual source models trained on Mapillary Vistas or BDD100K already contain diverse weather-related knowledge, UCDA further improves target-domain performance by selecting and consolidating reliable predictions from multiple source models.
Under the MiT-B5 setting, UCDA obtains the highest overall mIoU and achieves strong performance on safety-relevant categories such as road, sidewalk, car, and bus, confirming that the proposed collaborative adaptation strategy remains effective with Transformer-based perception models.
UCDA also improves categories that vary substantially with illumination and weather, such as sky and vegetation, suggesting that multi-source collaboration helps stabilize perception under environmental shifts frequently encountered in real driving scenarios.

\begin{figure*}[t!]
    \centering
    \includegraphics[width=0.999999\textwidth]{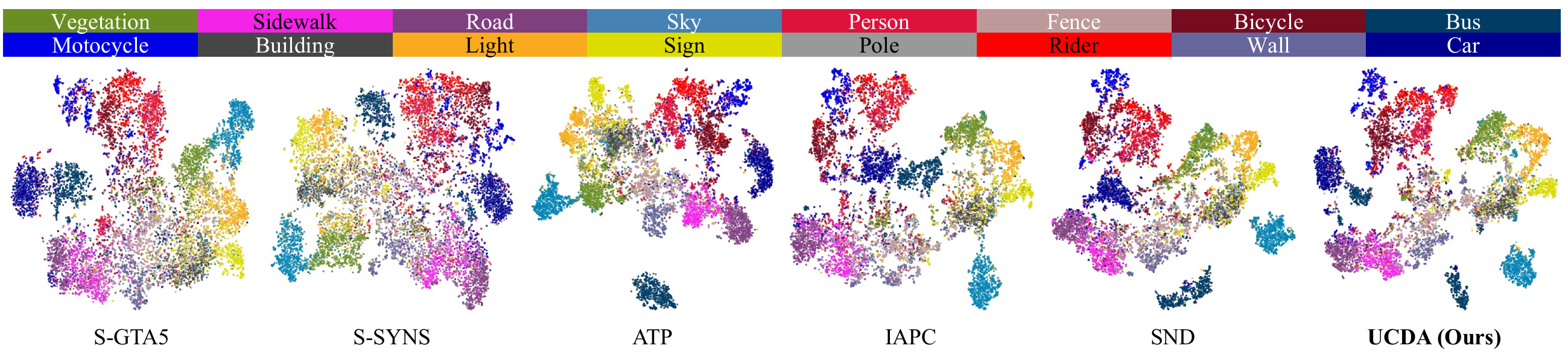}
    \caption{t-SNE visualization of target-domain feature distributions produced by different driving scene adaptation methods. The plots compare feature embeddings from source-only models, existing source-free methods including ATP, IAPC, and SND, and the proposed UCDA framework.}
    \label{fig.tsne}
\end{figure*}

Furthermore, the t-SNE visualization in Fig.~\ref{fig.tsne} compares target-domain feature embeddings on the Cityscapes 16-class task across source-only models, existing source-free methods, and UCDA.
UCDA forms more compact semantic clusters and clearer category boundaries, indicating that the collaborative adaptation process improves target-domain representation quality while preserving category-level discriminability for dense driving scene perception.

\subsection{Ablation Studies}
\label{sec.exp.ablation}

\begin{table}[t!]
    \tablesmallfont
    \centering
    \setlength{\tabcolsep}{0.55mm}
    \caption{Ablation analysis of the proposed UCDA framework on the Cityscapes 16-class protocol. ``CE'', ``POS'', and ``NEG'' denote cross-entropy loss, positive consistency loss, and negative consistency loss, respectively. mIoU$^{\ast}$ denotes the mean IoU calculated after excluding the challenging categories marked with $^{\ast}$.}
    \label{tab.each_part}
    \begin{tabular}{ccccccccc}
        \toprule
        \multicolumn{2}{c}{Source Models} & \multicolumn{3}{c}{Components} & \multicolumn{2}{c}{Collaborative Optimization} & \multicolumn{2}{c}{Knowledge Infusion} \\
        \cmidrule(r){1-2} \cmidrule(lr){3-5} \cmidrule(lr){6-7} \cmidrule(l){8-9}
        S-SYNS & S-GTA5 & CE & POS & NEG & mIoU (\%) & mIoU$^{\ast}$ (\%) & mIoU (\%) & mIoU$^{\ast}$ (\%) \\
        \midrule
        \checkmark &            &            &            &            & 37.2 & 42.9 & 50.2 & 55.7 \\ 
                   & \checkmark &            &            &            & 41.4 & 45.8 & 52.6 & 58.4 \\ 
        \midrule 
        \checkmark & \checkmark & \checkmark &            &            & 48.0 & 53.9 & 51.2 & 56.4 \\
        \checkmark & \checkmark & \checkmark & \checkmark &            & 50.5 & 55.6 & 52.8 & 57.9 \\
        \checkmark & \checkmark & \checkmark &            & \checkmark & 49.3 & 54.4 & 51.6 & 57.0 \\
        \checkmark & \checkmark & \checkmark & \checkmark & \checkmark & \textbf{52.6} & \textbf{58.4} & \textbf{54.6} & \textbf{60.2} \\
        \bottomrule
    \end{tabular}
\end{table}

Table~\ref{tab.each_part} summarizes the ablation study on the main components of UCDA across the two-stage adaptation process.
In the first stage, collaborative optimization refines the source models by exploiting complementary predictions from different source models.
In the second stage, knowledge infusion transfers the refined multi-source expertise into a unified target-domain model for deployment.
The results show that both stages contribute to target-domain performance, and knowledge infusion consistently improves the final target model over the collaboratively optimized source models.
Cross-entropy supervision from synthesized pseudo labels provides the primary semantic guidance and brings clear gains over the source-only baselines.
Adding the positive consistency loss further improves mIoU and mIoU$^{\ast}$ by encouraging alignment with reliable soft predictions in uncertain regions.
The negative consistency loss provides complementary suppression of unlikely categories, which helps reduce ambiguous predictions in difficult target-domain regions.
Combining positive and negative consistency losses achieves the best performance in both stages, confirming that explicit semantic guidance, positive consistency, and negative category suppression are all beneficial for source-free collaborative adaptation.

\subsection{Sensitivity Analysis of Key Hyperparameters}
\label{subsec:hyperparameters}

\begin{figure}[t!]
    \centering
    \includegraphics[width=0.48\textwidth]{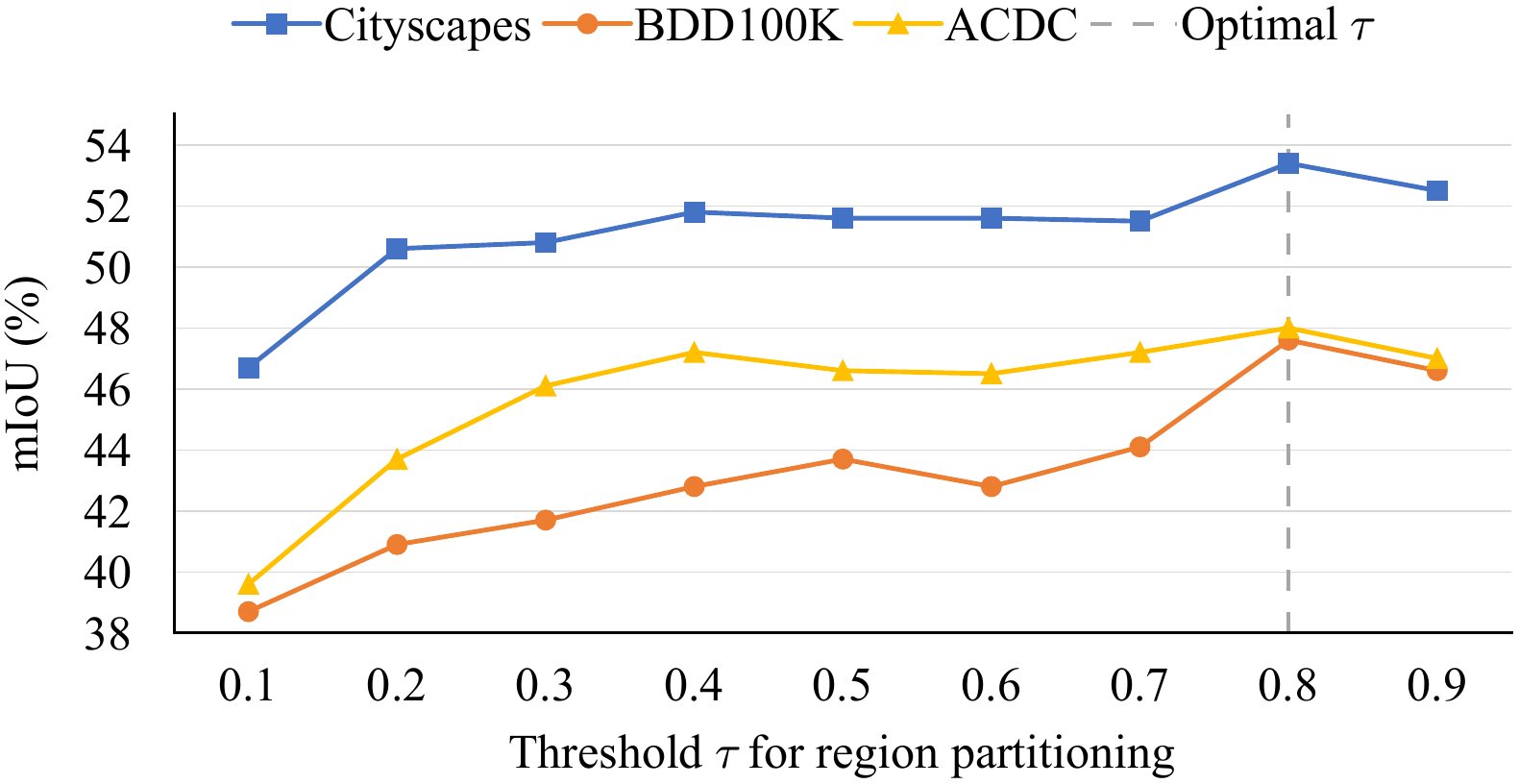}
    \caption{Quantitative sensitivity analysis of the confidence threshold $\tau$ for region partitioning on different datasets.}
    \label{fig.para_sensi}
\end{figure}

To evaluate the stability of UCDA with respect to region partitioning, a sensitivity analysis is conducted on the confidence threshold $\tau$.
As shown in Fig.~\ref{fig.para_sensi}, $\tau$ determines how target-domain pixels are divided between high-confidence regions supervised by synthesized pseudo labels and low-confidence regions regularized by positive and negative consistency constraints.
Across the evaluated datasets, UCDA achieves its best performance when $\tau$ is set to $0.8$.
This setting applies discrete semantic supervision only to sufficiently confident predictions, reducing the risk of propagating noisy pseudo labels from source models.
At the same time, uncertain regions remain available for soft positive and negative consistency regularization, which helps stabilize adaptation in ambiguous target-domain areas.
The relatively smooth performance variation around the optimal threshold indicates that UCDA is not overly sensitive to this hyperparameter, supporting its robustness across different target driving domains.
In all experiments, the loss weights are fixed to isolate the effect of the region-partitioning threshold.
Specifically, the weights for cross-entropy loss (\ie $\lambda_{ce}$ and $\lambda_{ce}^k$), positive consistency loss (\ie $\lambda_p$ and $\lambda_p^k$), and negative consistency loss (\ie $\lambda_n$ and $\lambda_n^k$) are set to $0.8$, $0.1$, and $0.1$, respectively.
This configuration keeps synthesized pseudo labels as the primary semantic guidance while using the two consistency losses as auxiliary regularization.
The larger cross-entropy weight emphasizes explicit supervision in high-confidence regions.
The smaller positive and negative consistency weights regularize uncertain regions without overwhelming the pseudo-label supervision.
This fixed weighting scheme enables the sensitivity analysis to focus on how the threshold $\tau$ balances explicit supervision and soft regularization.

\subsection{Evaluation of Prototype-Aware Pseudo-Label Synthesis}
\label{sec.exp.Evaluation_of_cpb}

\begin{table}[t!]
    \centering
    \tablesmallfont
    \setlength{\tabcolsep}{3.1mm}
    \caption{Evaluation of synthesized pseudo-label quality on the Cityscapes 16-class protocol. The table compares source-only predictions, baseline fusion strategies, and UCDA variants with Euclidean (E) or cosine (C) reliability estimation. Checkmarks indicate the source models used for pseudo-label synthesis.}
    \label{tab.ps_eval}
    \begin{tabular}{c|ccc|c}
        \toprule
        \multirow{2}{*}{Method} & \multicolumn{3}{c|}{Source Models} & \multirow{2}{*}{mIoU (\%)} \\
        \cmidrule(lr){2-4}
        & S-SYNS & S-SYNC & S-GTA5 & \\
        \midrule
        S-SYNS~\cite{synthia_data} & \checkmark &  &  & 37.2 \\
        S-SYNC~\cite{synscapes} &  & \checkmark &  & 51.4 \\
        S-GTA5~\cite{gta_dataset} &  &  & \checkmark & 43.4 \\
        \midrule
        Majority Vote & \checkmark & \checkmark & \checkmark & 49.9 \\
        Probability Average & \checkmark & \checkmark & \checkmark & 52.8 \\
        Entropy~\cite{wang2024curriculum} & \checkmark & \checkmark & \checkmark & 51.7 \\
        Softmax~\cite{Tiandbc} & \checkmark & \checkmark & \checkmark & 51.8 \\
        \midrule
        \multirow{4}{*}{\textbf{UCDA (E)}}
        & \checkmark &  & \checkmark & 47.5 \\
        &  & \checkmark & \checkmark & 52.3 \\
        & \checkmark & \checkmark &  & 49.9 \\
        & \checkmark & \checkmark & \checkmark & 52.1 \\
        \midrule
        \multirow{4}{*}{\textbf{UCDA (C)}}
        & \checkmark &  & \checkmark & 47.4 \\
        & \checkmark & \checkmark &  & 50.7 \\
        &  & \checkmark & \checkmark & 53.1 \\
        & \checkmark & \checkmark & \checkmark & \textbf{53.7} \\
        \bottomrule
    \end{tabular}
\end{table}

\begin{figure*}[t!]
    \centering
    \includegraphics[width=0.999999\textwidth]{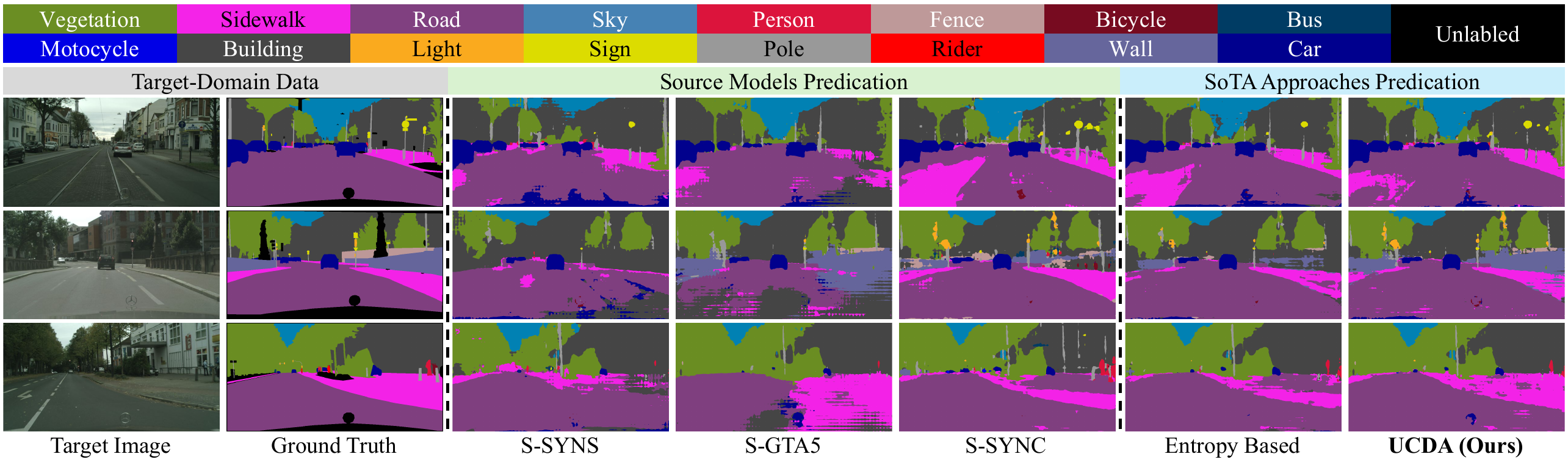}
    \vspace{-18pt}
    \caption{Visual comparison of synthesized pseudo labels generated by various approaches for the 16-class parsing task on the Cityscapes dataset.}
    \label{fig.visual_ps_label}
\end{figure*}

Table~\ref{tab.ps_eval} summarizes the pseudo-label quality evaluation on the Cityscapes 16-class protocol, while Fig.~\ref{fig.visual_ps_label} provides the corresponding visual comparison.
Compared with source-only predictions and baseline fusion strategies, the prototype-aware synthesis used in UCDA improves pseudo-label accuracy by estimating prediction reliability through class-level prototypes.
The cosine-based variant achieves the best mIoU, outperforming the Euclidean variant and conventional entropy- or softmax-based selection.
Since the source models are trained on different domains, their prediction distributions and confidence scales are not directly comparable.
Cosine similarity is less affected by magnitude differences across model-specific logits and therefore provides a more stable criterion for selecting reliable pseudo labels.
The results also show that using complementary source models improves pseudo-label quality compared with relying on a single source model, indicating that source diversity provides useful supervision for target-domain perception.
The visual results in Fig.~\ref{fig.visual_ps_label} further support this observation, showing that UCDA produces cleaner pseudo-label maps with fewer fragmented or inconsistent regions than the baseline strategies.
These improvements are particularly important for visually ambiguous regions in driving scenes, where unreliable pseudo labels can compromise target-domain perception during source-free adaptation.
Overall, the quantitative and visual comparisons demonstrate that prototype-aware reliability estimation provides more trustworthy pseudo labels, enabling UCDA to integrate complementary source knowledge more effectively for target-domain dense perception.

\subsection{Generalization Across Target Driving Domains}
\label{sec.exp.gene}

\begin{table}[t!]
	\setlength{\heavyrulewidth}{2pt}
    \tablesmallfont
	\centering
	\setlength{\tabcolsep}{3.5mm}
    \caption{Generalization performance comparison across different real-world target driving domains. The average mIoU summarizes the overall cross-domain robustness of each method.}
	\begin{tabular}
		{c|ccc|c}
		\toprule[0.7pt]
        Method &Cityscapes &BDD100K &ACDC &Average \\
        \midrule
        S-GTA5 &38.7 &34.5 &27.6 &33.6\\
        \midrule
        IAPC~\cite{Caoiapc} &48.5 &41.3 &35.7 &41.8\\
        SND~\cite{SND} &53.3 &33.5 &36.0 &40.9 \\
        ATP~\cite{wang2024curriculum} &46.2 &32.6 &28.8 &36.0\\
        \midrule
        \textbf{UCDA (Ours)} & \textbf{60.4} & \textbf{45.7} & \textbf{39.3} & \textbf{48.5}\\
        \bottomrule[0.7pt]
	\end{tabular}
    \label{tab.gene}
\end{table}

The preceding experiments evaluate adaptation performance on individual target domains, where the training and evaluation data are drawn from the same target benchmark. However, autonomous driving systems may encounter deployment environments that differ from the target data used during adaptation.
To examine whether the adapted model can maintain reliable perception under such target-domain changes, cross-domain generalization experiments are conducted across Cityscapes, BDD100K, and ACDC.
Specifically, all methods are adapted on Cityscapes as the unlabeled target domain and then evaluated on Cityscapes, BDD100K, and ACDC without further adaptation.
In this setting, the Cityscapes results reflect in-domain target performance, whereas the BDD100K and ACDC results assess cross-target generalization to unseen real-world driving conditions with different scene layouts, weather, and illumination.
The average mIoU over the three datasets is reported to summarize the overall generalization ability of each method.
As reported in Table~\ref{tab.gene}, UCDA achieves the best performance on all evaluated datasets after being adapted only on Cityscapes.
The improvement is not limited to the in-domain Cityscapes evaluation; UCDA also retains clear advantages on BDD100K and ACDC, indicating stronger cross-target generalization than single-model adaptation methods.
This result suggests that integrating complementary knowledge from multiple source models reduces dependence on a single target-domain distribution and improves the robustness of dense perception under unseen driving conditions.
Such generalization is important for autonomous driving systems, where the deployment environment may differ from the data available during offline adaptation.

\subsection{Real-World Evaluation on an Autonomous Vehicle Platform}
\label{sec.exp.NIO_data}

\begin{table}[t!]
	\setlength{\heavyrulewidth}{2pt}
    \tablesmallfont
	\centering
	\setlength{\tabcolsep}{1.6mm}
    \caption{Comparison of deployment efficiency and parsing accuracy on the NIO dataset.}
	\begin{tabular}
		{c|ccccc}
		\toprule[0.7pt]
        Method &Params (M) &FLOPs (G) &FPS &Latency (ms) &mIoU ($\%$) \\
        \midrule
        S-GTA5 &43.9 &731.9 &51.8 &19.3 &35.8\\
        S-BDD &43.9 &731.9 &51.3 &19.5 &51.9\\
        \midrule
        IAPC~\cite{Caoiapc} &43.9 &731.9 &51.0 &19.6 &50.5\\
        SND~\cite{SND} &43.9 &731.9 &50.8 &19.7 &52.6\\
        ATP~\cite{wang2024curriculum} &43.9 &731.9 &50.5 &19.8 &51.3\\
        \midrule
        \textbf{UCDA (Ours)} &43.9 &731.9 &51.3 &19.5 &55.8\\
        \bottomrule[0.7pt]
	\end{tabular}
    \label{tab.deployment_efficiency}
\end{table}

\begin{figure}[t!]
    \centering
    \includegraphics[width=0.46\textwidth]{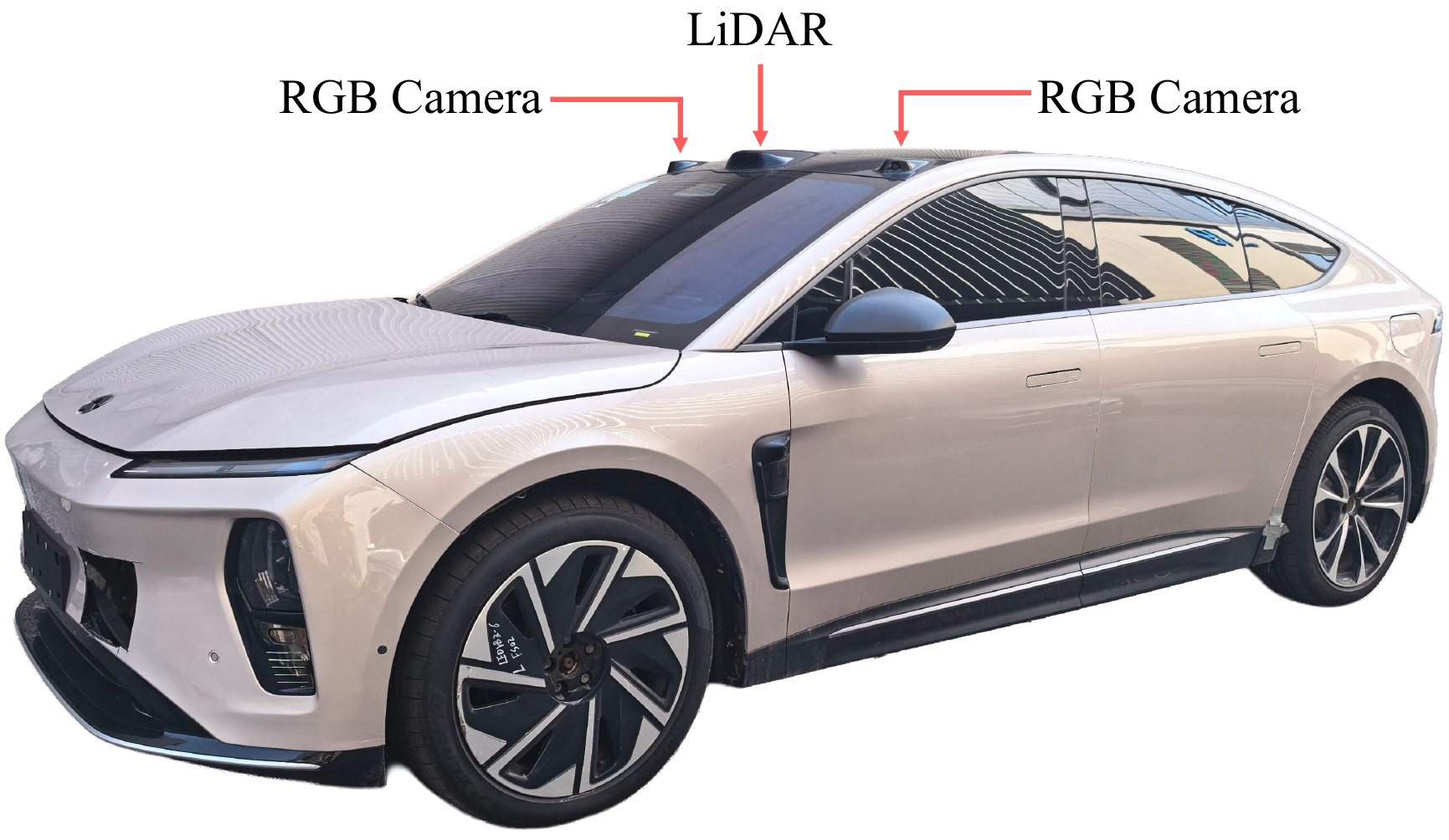}
    \caption{The NIO ET9 vehicle utilized as the experimental platform for real-world driving data acquisition. This vehicle serves as the mobile sensing unit to capture diverse urban scenarios, providing the target-domain data required to evaluate the robustness of our proposed UCDA framework in practical autonomous driving environments.}
    \label{fig.car}
\end{figure}

\begin{figure*}[t]
    \centering
    \includegraphics[width=0.999999\textwidth]{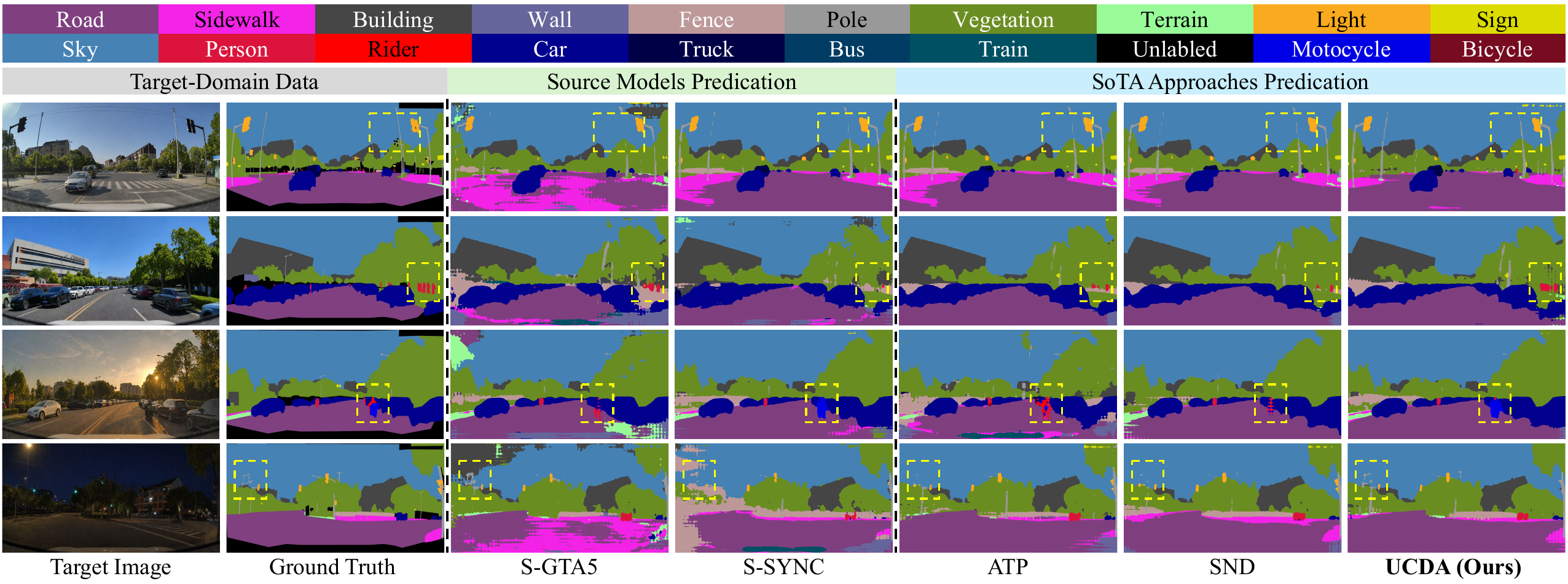}
    \vspace{-18pt} 
    \caption{Qualitative results on the NIO dataset collected in real-world scenarios.}
    \label{fig.NIO19}
\end{figure*}

To assess the applicability of UCDA in realistic autonomous driving environments beyond public benchmarks, real-world experiments are conducted using the NIO ET9 vehicle platform shown in Fig.~\ref{fig.car}.
The platform is equipped with 33 high-precision sensors, including a long-range LiDAR and seven high-definition cameras with a resolution of $8\times10^{6}$ pixels each, enabling multi-view data acquisition in real traffic environments.
The collected data cover diverse urban scenarios with complex traffic conditions and varied road layouts.
This setting provides a practical evaluation of whether UCDA can maintain reliable dense perception beyond standard benchmark datasets~\cite{cityscape,bdd100,acdc_data}.
As shown in Fig.~\ref{fig.NIO19}, UCDA produces coherent parsing results across 16 semantic categories in real-world driving scenes.
The quantitative results in Table~\ref{tab.deployment_efficiency} further show that UCDA improves mIoU over source-only baselines and existing source-free adaptation methods while using the same deployed target-model architecture.
Since all methods share the same inference model size and computational cost, the accuracy gain of UCDA does not introduce additional deployment overhead.

\section{Discussion}
\label{sec.discussion}



\subsection{Extension to Diversified Multi-Source Configurations}
\label{sec.dis.Multi-Source_Configurations}


\begin{table}[t!]
	\setlength{\heavyrulewidth}{2pt}
    \tablesmallfont
	\centering
	\setlength{\tabcolsep}{1.5mm}
    \caption{Performance of UCDA under different source-model configurations on the Cityscapes 16-class protocol. mIoU$^{\ast}$ denotes the mean IoU calculated after excluding the challenging categories marked with $^{\ast}$.}
	\begin{tabular}{ccccccc}
		\toprule[0.7pt]
        \multicolumn{3}{c}{Synthetic Sources} & \multicolumn{2}{c}{Real-World Sources} & \multicolumn{2}{c}{Metrics} \\
        \cmidrule(lr){1-3} \cmidrule(lr){4-5} \cmidrule(lr){6-7}
        S-SYNS & S-GTA5 & S-SYNC & S-BDD & S-MAP & mIoU (\%) & mIoU$^{\ast}$ (\%) \\
		\midrule
		\checkmark & & & & &37.2 &42.9\\
  	    & \checkmark & & & &41.4 &45.8\\
        & & \checkmark & & &51.6 &56.8\\
        & & &\checkmark & &58.4 &63.9\\
        & & & & \checkmark &65.0 &69.8\\
        \midrule
        \checkmark &\checkmark & & & &52.8 &58.5 \\
        \checkmark & &\checkmark & & &54.6 &60.2\\
        \checkmark  & & &\checkmark & &55.4 &61.2\\
        & & & \checkmark  &\checkmark &65.5	&70.5\\
        \checkmark &\checkmark &\checkmark & & &56.5 &62.0\\
        \checkmark &\checkmark &\checkmark &\checkmark & &57.7 &63.2\\
        \checkmark &\checkmark &\checkmark &\checkmark &\checkmark &\textbf{66.1} &\textbf{71.5}\\
        \bottomrule[0.7pt]
	\end{tabular}
	\label{tab.multi_source}
\end{table}

To assess the scalability of UCDA under different source-model configurations, Table~\ref{tab.multi_source} reports the results on the Cityscapes 16-class protocol.
The results show that combining multiple source models generally improves target-domain performance over individual source models, indicating the benefit of exploiting complementary source-domain expertise.
Notably, combinations of synthetic source models already bring clear gains, while incorporating real-world source models further improves performance.
These observations suggest that UCDA can exploit source models with different visual and semantic biases, allowing their complementary predictions to support more reliable target-domain perception.
When all five source models are used, the source domains become more heterogeneous in terms of data origin, scene characteristics, and prediction behavior.
Despite this increased heterogeneity, UCDA achieves the best mIoU and mIoU$^{\ast}$, demonstrating its ability to consolidate diverse source expertise for reliable dense perception in the target driving domain.

\subsection{Limitations}
\label{sec.dis.limitations}

Despite the demonstrated effectiveness of UCDA, several limitations remain.
The effectiveness of UCDA depends on the quality and diversity of the available source models, since weak or highly biased source models may provide limited complementary information.
The absence of target-domain annotations also makes it difficult to completely filter noisy predictions from different source models.
Although prototype-aware reliability estimation mitigates this issue, incorrect predictions may still be selected when they align strongly with an incorrect class prototype.
Such cases can mislead collaborative optimization and propagate errors to the final target model.
This limitation may lead to reduced performance on specific categories, as observed for the road category in Table~\ref{tab.cityscapes_19}.
In addition, using more source models increases the training-time computational and memory cost, although the deployed target model retains the same inference architecture.

\subsection{Potential Applications}
\label{sec.dis.applications}

Beyond autonomous driving, UCDA can be extended to privacy-sensitive perception scenarios where raw data cannot be freely shared.
Because it consolidates knowledge from multiple pre-trained models without accessing the original source data, UCDA is suitable for cross-organization and cross-region adaptation.
Potential applications include collaborative vehicle fleets in intelligent transportation, medical image analysis across hospitals, and security monitoring across cities, where privacy, regulation, or ownership constraints may prevent centralized data sharing.
In these scenarios, the UCDA paradigm provides a practical way to improve target-domain perception while reducing dependence on direct data exchange.

\section{Conclusion and Future Work}
\label{sec.conclusion}

This article presented an unsupervised collaborative domain adaptation framework for driving scene parsing, aiming to overcome the source-specific limitations of single-model adaptation.
The framework estimates cross-model prediction reliability in a source-free setting, enabling complementary knowledge to be selected from multiple pre-trained source models without accessing their original training data.
A class-level prototype memory bank provides semantic references for reliability estimation, reducing the influence of inconsistent confidence scales across independently trained models.
A two-stage adaptation strategy further refines source models through collaborative optimization and then distills their validated expertise into a unified target-domain model.
Extensive evaluations on synthetic-to-real and real-to-real adaptation tasks, together with data collected from an autonomous vehicle platform, demonstrate that UCDA improves target-domain dense perception under diverse driving conditions.
The results show that multi-source collaboration improves challenging semantic categories, enhances cross-target generalization, and preserves deployment efficiency by using a single target model during inference.
Future work will explore the integration of additional sensing modalities and the extension of UCDA toward real-time adaptation in dynamic driving environments.

\bibliographystyle{IEEEtran}
\bibliography{refs}

\vfill

\end{document}